%% file: main.tex
\documentclass[acmlarge,nonacm,screen]{acmart}

\usepackage{wrapfig}
\usepackage{xspace}
\usepackage{float}     
\usepackage{booktabs}
\usepackage{placeins}  
\usepackage{subcaption}
\usepackage{graphicx}
\usepackage{pifont}
\usepackage{subcaption}
\usepackage{makecell}
\usepackage{svg}
\usepackage{enumitem}
\usepackage[ruled,linesnumbered,vlined]{algorithm2e}
\usepackage{colortbl}
\usepackage{multirow}
\usepackage[utf8]{inputenc}
\usepackage[normalem]{ulem}  
\usepackage{soul}            
\usepackage[dvipsnames]{xcolor}
\definecolor{oursblue}{HTML}{1F4E99}

\setcopyright{none}

\newcommand{\revmode}{3}        
\definecolor{revdel}{RGB}{190,30,30}    
\definecolor{revnew}{RGB}{0,0,160}      

\ifnum\revmode=1
  \newcommand{\old}[1]{{\color{revdel}\sout{#1}}}
  
\fi
\ifnum\revmode=2
  \newcommand{\old}[1]{}
  
\fi
\ifnum\revmode=3
  \newcommand{\old}[1]{}
  
\fi

\newenvironment{tightitemize}{
  \begin{itemize}[nosep, leftmargin=*]
}{\end{itemize}}

\newcommand{\name}{\textit{\textbf{HyVox}}\xspace}

\newif\ifhighlight
\highlightfalse    

\definecolor{lightblue}{RGB}{200,230,255}
\definecolor{lightgreen}{RGB}{210,255,210}
\definecolor{lightyellow}{RGB}{255,250,200}
\definecolor{lightpink}{RGB}{255,220,230}
\definecolor{lightgraycustom}{RGB}{230,230,240}

\begin{document}

\title{\name: Rethinking Low-Power Voice Intelligence with Hybrid Neuromorphic–von Neumann Edge Computing}

\author{\href{https://orcid.org/0000-0001-8807-997X}{Snehasis Addy}}
\affiliation{%
  \institution{University of Massachusetts Amherst}
  \city{Amherst}
  \state{MA}
  \country{USA}}
\email{saddy@umass.edu}

\author{\href{https://orcid.org/}{Eduardo Calle Ortiz}}
\affiliation{%
  \institution{University of Massachusetts Amherst}
  \city{Amherst}
  \state{MA}
  \country{USA}}
\email{ecalleortiz@umass.edu}

\author{\href{https://orcid.org/}{Prashant Shenoy}}
\authornote{Both authors are corresponding authors.}
\affiliation{%
  \institution{University of Massachusetts Amherst}
  \city{Amherst}
  \state{MA}
  \country{USA}}
\email{shenoy@umass.edu}

\author{\href{https://orcid.org/0000-0001-8078-4463}{Phuc Nguyen}}
\authornotemark[1]
\affiliation{%
  \institution{University of Massachusetts Amherst}
  \city{Amherst}
  \state{MA}
  \country{USA}}
\email{vp.nguyen@umass.edu}

\renewcommand{\shortauthors}{Addy et al.}

\begin{abstract}

In this paper, we present \name, a hybrid neuromorphic--von Neumann system for
low-energy open-set speaker identification under \textit{extremely tight
energy budgets}. \name splits the workload at the speaker embedding. A
\textit{neuromorphic processor} embeds the speech with
sparse, event-driven computation, and a \textit{von Neumann microcontroller}
holds the speaker gallery and makes every decision: it scores queries by
cosine similarity, rejects unknown speakers, and enrolls new users by writing
a prototype, with no weight update. On the neuromorphic processor, a
hierarchical temporal abstraction (HTA) SNN backbone learns a synaptic time
constant per leaky integrate-and-fire (LIF) neuron, allocating $832$ neurons
across fast, medium, and slow timescales. A $32$-neuron spiking embedding
extractor (SEE), pretrained on episodes that each enroll new speakers and
reject outsiders, turns the backbone output into an embedding for speakers
unseen in training; its $32$ spike counts are the only interface between the
two processors. To fit the chip's synapse and integer budgets, we compress
the backbone $6.5\times$ to $29.0$k synapses, train it for integer bit-shift
dynamics, and fine-tune it with chip-accurate spiking training (CAST). CAST
trains through a differentiable replica of the chip's integer neuron update,
derived from its cycle-accurate simulator, so that gradient descent optimizes
the exact weights and thresholds the chip runs; it doubles the top-1 accuracy
of the mapped backbone on unseen speakers. On $20$ Common Voice households of
unseen speakers, \name reaches $99.6\%$ top-1 and $5.5\%$ rejection EER with
$2$ members and $94.9\%$ and $13.2\%$ with $20$, and the network deployed on
the neuromorphic chip matches its trained model. Each inference on a
$3$-s window takes $0.96\,\mathrm{s}$ and $1.63\,\mathrm{mJ}$ of dynamic energy, $42\%$
lower latency than the fastest embedded ANN baseline and $13.5\times$ less
dynamic and $5.8\times$ less total energy than the most energy-efficient one.

\end{abstract}

\maketitle
\renewcommand{\shortauthors}{Addy et al.}

\input{Sections/1-introduction_new}
\input{Sections/2-background_new}
\input{Sections/3-system_overview_new}
\input{Sections/4-technique1}

\input{Sections/5-technique2_SEE}

\input{Sections/6-technique3_new}

\input{Sections/7-evaluation_new}
\input{Sections/8-related_work}

\input{Sections/9-discussion}

\input{Sections/10-conlclusion}

\bibliographystyle{ACM-Reference-Format}
\bibliography{ref-1.bib}
\end{document}

%% file: Sections/1-introduction_new.tex
\section{Introduction}
\label{sec:introduction}

\textbf{Motivation.}
Smart audio devices are moving from fixed command
recognition~\cite{warden2018speechcommands,zhang2018helloedge,kim2021bcresnet}
toward personalized speaker
identity~\cite{variani2014dvector,snyder2018xvectors,gallacher2023speakerid}.
A shared smart speaker adds household members after it is installed, a
hearable recognizes its owner, and a home appliance tailors its responses to
the enrolled user who speaks and treats strangers as guests. In each case the speakers that matter are
defined after deployment~\cite{snell2017prototypical,wan2018ge2e}: the device
enrolls a new user from a short recording, keeps that user's representation in
local memory, compares every query against a gallery that changes over time,
and rejects speech from anyone who is not
enrolled~\cite{scheirer2013open,geng2021opensetsurvey,vaze2022openset}.
These operations run on battery-powered hardware, so the energy of the
inference that turns a window of speech into speaker evidence bounds how many
decisions the device can make on one
charge~\cite{david2021tflmicro,banbury2021mlperf,yik2025neurobench}.

\begin{figure}[t]
  \centering
  \includegraphics[width=0.9\linewidth]{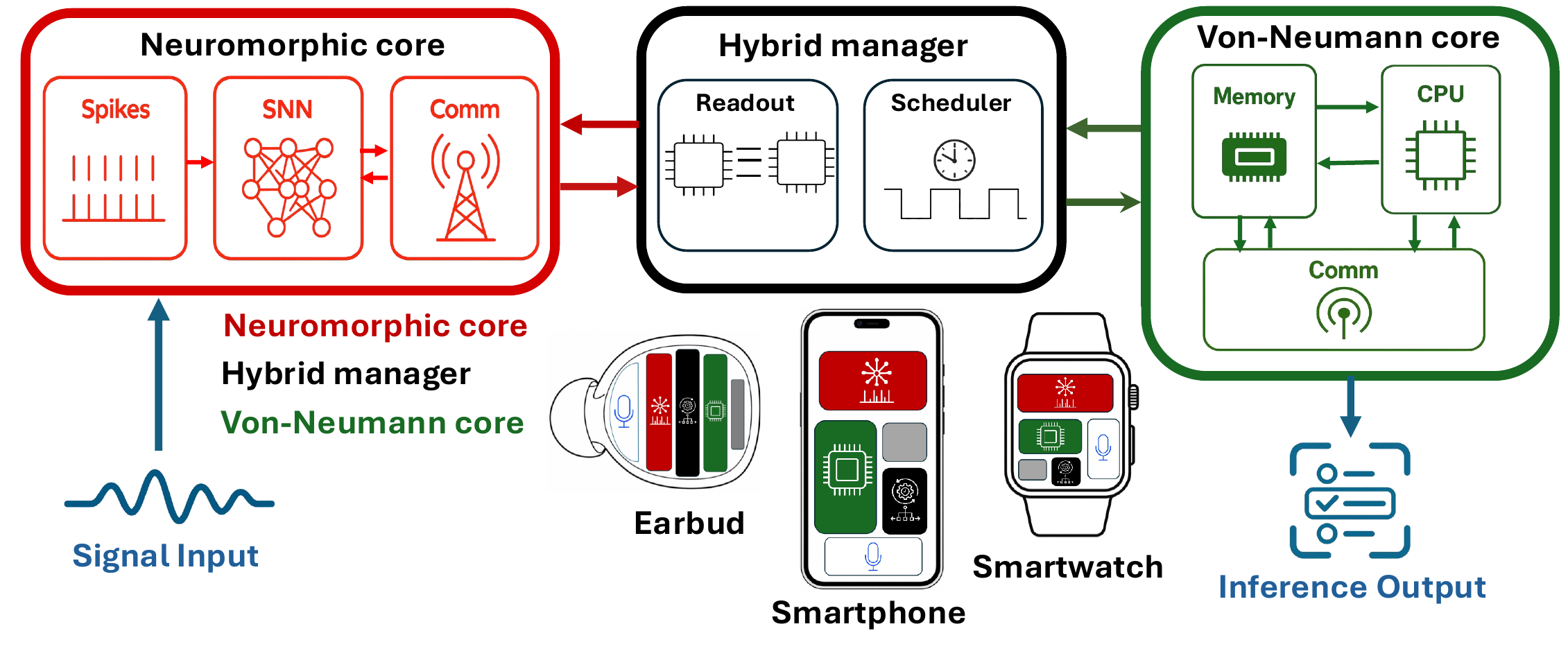}
  \vspace{-6pt}
  \caption{\name's concept: Hybrid platform for low-energy speaker identification}
  \vspace{-18pt}
  \label{fig:hyvox-concept}
\end{figure}

\noindent \textbf{Current Approaches and Limitations.} Speaker
identification on edge devices runs as a TinyML pipeline on a von Neumann
microcontroller. A compact artificial neural network (ANN), such as
DS-CNN~\cite{zhang2018helloedge}, BC-ResNet~\cite{kim2021bcresnet}, or the
small-footprint speaker network of Balian et
al.~\cite{balian2021smallfootprint}, embeds each window of audio, and the
same microcontroller scores the embedding against a gallery of enrolled
prototypes, rejects unknown speakers, and enrolls new users. Scoring takes
only a few dot products per enrolled speaker, so ANN inference accounts for
nearly all of the energy of a decision. On a von Neumann core, each
multiply--accumulate of this inference fetches its weight and activation from
memory, so this data movement sets the energy of every inference.
Neuromorphic processors such as Xylo Audio~3~\cite{synsense2023xyloaudio3ds} remove this
traffic: an event-driven front end and spiking cores keep weights beside the
neurons and compute only when spikes arrive. A neuromorphic core running a
spiking neural network (SNN) is therefore a natural replacement for the ANN
in the TinyML pipeline. Moving the entire open-set pipeline onto the
neuromorphic core, however, is infeasible. The core executes only the SNN
mapped onto it and has no general-purpose compute, so it cannot store a
gallery that changes at runtime, normalize and compare embeddings, apply a
rejection threshold, or enroll a speaker without reconfiguring the network.
Existing neuromorphic audio work~\cite{muir2023xylo,bos2024micropower}
accordingly targets closed-set classification and keyword spotting.
The same co-location that saves energy also caps the network size: every
weight occupies on-chip memory beside its neuron, so Xylo Audio~3 holds at
most $1{,}000$ spiking neurons and $31{,}744$ int8 recurrent synapses, orders
of magnitude below the millions of parameters of standard speaker
encoders~\cite{snyder2018xvectors}.
Two questions remain open: how to learn a speaker embedding that generalizes
to unseen speakers within this fixed hardware budget, and how to design a system that combines the low-energy inference of a
neuromorphic chip with the general-purpose compute that open-set recognition
requires.

\noindent \textbf{Proposed Approach.} In this paper, we present \name, a
hybrid neuromorphic--von Neumann system for low-energy open-set speaker identification
(Fig.~\ref{fig:hyvox-concept}). The central idea is to split the workload at
the embedding: the neuromorphic core takes over the ANN's role in the TinyML
pipeline, and the von Neumann microcontroller (MCU) keeps every operation that needs
general-purpose compute. Xylo Audio~3 runs the audio front end (AFE) and a spiking embedding
network and emits one 32-component spike-count vector per inference on a
three-second window. An STM32L476 MCU host turns these
counts into decisions: it holds a gallery of speaker prototypes, scores each
query by cosine similarity, rejects unknown speakers, and enrolls new users by
writing a prototype. The spiking network is shared by every device and never
trained on a user's audio; each household differs only in the gallery on its
host.

\noindent \textbf{Challenges.}
Moving the embedding from an ANN on the MCU to an SNN on the neuromorphic core
places the whole burden of recognition on a network of fixed size, integer
dynamics, and a $32$-neuron output, and that output must still support every
open-set decision the MCU makes. This raises four challenges, from the design
of the network to its execution on the chip:
\begin{tightitemize}
    \item \textbf{Speaker evidence in a small SNN.} Speaker evidence
    spans milliseconds (formant transitions, plosive releases) to hundreds of
    milliseconds (pitch contour, speaking rate). ANN encoders cover this range
    with depth and width, but the chip holds $1{,}000$ neurons. In a spiking
    network the only memory of the past is each neuron's decay, so the
    timescales assigned to so few neurons determine which speaker evidence the
    network can represent at all.
    \item \textbf{A $32$-dimensional spiking embedding for unseen speakers.}
    The chip's hidden state never leaves the chip; the MCU receives only the
    spike count of each of the $32$ readout neurons per window. These counts
    are the entire interface, so they must serve directly as the embedding:
    they must identify speakers absent from training, enroll a new speaker
    from one minute of speech with no weight update, and separate speakers who were never enrolled so the MCU can reject them. Large-corpus pretraining
    separates a fixed set of training speakers and never practises enrollment
    or rejection.
    \item \textbf{Fitting the hardware's synapse and precision budget.}
     On-chip memory bounds the number of synapses; Xylo Audio~3 holds at most
    $31{,}744$. A fully connected network of five layers of width $w$ has
    four connections between adjacent layers of $w^2$ synapses each, so the
    budget allows
    $w\approx\sqrt{31{,}744/4}\approx 89$, about $445$ neurons in total. That
    is under half of Xylo's $1{,}000$ and too few to cover the timescales of
    speaker evidence. A network with enough neurons exceeds the
    synapse budget several times over and must lose most of its synapses to
    fit, and the hardware further restricts weights to low-precision integers
    and decays to a discrete set, int8 weights and power-of-two bit shifts on
    Xylo. Each of these steps removes speaker information that training must
    recover.
    \item \textbf{The gap between the trained network and the chip.} Even
    after compression and quantization, the chip computes differently from
    the model used in training. On the chip, a neuron can fire only a limited
    number of times per timestep, signals take extra timesteps to pass from
    one layer to the next, and neuron state is clipped when it exceeds the
    register range. Training ignores all three, so the network learns to
    rely on behavior the chip cannot reproduce, and a network that recognizes
    speakers well in training can lose much of its accuracy once it runs on
    the chip.
\end{tightitemize}

\noindent \textbf{Technical Contributions.} \name addresses these challenges
with four matching contributions:
\begin{tightitemize}
    \item \textbf{HTA-SNN backbone.}  We design a hierarchical temporal
    abstraction (HTA) SNN backbone of $832$ leaky integrate-and-fire (LIF) neurons that learns a
    synaptic time constant per neuron, so training allocates the neuron
    budget across timescales, with cross-scale projections that carry fast
    acoustic evidence to the slow stages (Sec.~\ref{sec:hta-architecture}).
    At the same synapse budget, it exceeds SynNet~\cite{muir2023xylo,bos2024micropower}, the reference SNN for
    Xylo, by $4.1$ top-1 points and $1.6$ equal error rate (EER) points.
    \item \textbf{Spiking embedding extractor (SEE).} We propose a
    $32$-neuron spiking readout pretrained episodically on speakers disjoint
    from the backbone's: every episode enrolls new speakers from a few
    windows and rejects speakers who are not enrolled (Sec.~\ref{sec:see}).
    Its normalized spike counts are the trained embedding, so the STM32 host
    holds every speaker-specific quantity, the gallery, a sliding
    three-window buffer, and a threshold per gallery size, enrolls a user
    with forward passes and a prototype write, and decides in under $10^3$
    operations with $64$ bytes per speaker, one signed 16-bit value per
    embedding dimension (Sec.~\ref{sec:host-inference}).
    \item \textbf{Dense-to-sparse compression.} Since a large sparse network
    outperforms a small dense one with the same number of
    weights~\cite{zhu2017prune}, we train the backbone dense, with $188.4$k
    synapses ($5.9\times$ the budget), and compress it $6.5\times$ to
    $29.0$k synapses, $15.4\%$ of the dense network, on a $293$-timestep
    task. To achieve this, we extend SpikeFit~\cite{spikefit2025}, whose
    Fisher spike contribution scores channels and was demonstrated on a
    $4$-timestep image classifier at a single pruning ratio, to a
    per-synapse score ranked globally across stages and skip connections,
    and prune iteratively in $14$ steps while the learned time constants
    settle onto the chip's bit-shift grid (Sec.~\ref{sec:compression}).
    Training the compressed network
    for the chip's integer bitshift dynamics then recovers its accuracy
    before quantization (Secs.~\ref{sec:bitshift-qat}
    and~\ref{sec:eval-ondevice}).
    \item \textbf{Chip-accurate spiking training (CAST).} We derive a
    trainable PyTorch replica of the chip's integer neuron update from the
    chip's cycle-accurate simulator and fine-tune the network through it, a
    method that applies to any digital neuromorphic chip with such a
    simulator. Our CAST replica for Xylo reproduces $99.8\%$ of the
    simulator's hidden spike counts
    exactly. Its trainable parameters are the chip's own int8 weights and
    integer thresholds, so gradient descent optimizes the configuration the
    chip runs, under its event limits and per-layer delay. CAST doubles the top-1 of the mapped backbone on unseen
    speakers, from $31\%$ to $61\%$ (Sec.~\ref{sec:eval-ondevice}).
\end{tightitemize}

\noindent We evaluate \name on $20$ accent-homogeneous Common Voice households of
speakers unseen by both the backbone and the SEE. \name reaches $99.6\%$
top-1 and $5.5\%$ rejection EER with $2$ members and $94.9\%$ and $13.2\%$
with $20$. On $20$ further households, the deployed system on the chip matches or exceeds the
PyTorch model it was fine-tuned from, with $96.7\%$ top-1 and $10.0\%$ EER at $n=20$. \name
reduces the inference energy. Each inference on a three-second window takes
$0.96\,\mathrm{s}$ and $1.63\,\mathrm{mJ}$ of dynamic energy, $13.5\times$
less dynamic and $5.8\times$ less total energy than the most energy-efficient
embedded ANN baseline, with a $42\%$ lower latency than the fastest one.

%% file: Sections/2-background_new.tex
\section{Background}
\label{sec:background}

This section covers speaker embeddings and how TinyML pipelines compute them
on embedded devices, spiking neural networks (SNNs) and the neuromorphic
processors that run them, and how a trained SNN is fitted to neuromorphic
hardware.

\textbf{Speaker embeddings and the TinyML pipeline.}
Speaker recognition is commonly framed as an embedding problem: an encoder
maps each utterance to a vector whose geometry places same-speaker examples
close together and different-speaker examples far
apart~\cite{variani2014dvector,snyder2018xvectors}. Angular-margin
classification~\cite{deng2019arcface} learns this geometry from a large set of
labelled training speakers, and supervised contrastive
learning~\cite{khosla2020supcon} shapes it directly over pairs of examples. In
open-set recognition the deployed speakers are not known at training
time~\cite{geng2021opensetsurvey,vaze2022openset}: the device enrolls each
user as a normalized prototype~\cite{snell2017prototypical}, scores a query
against every prototype by cosine similarity, and rejects the query when its
best score falls below a threshold. Few-shot open-set
learning~\cite{liu2020peeler,kc2022openfeat} trains for this use. Each
training episode samples a new set of speakers, builds their prototypes from a
few examples, classifies further examples of them, and includes speakers
outside the set that must be rejected, so the model practises enrollment and
rejection on speakers that change in every episode.
On an edge device, the whole pipeline runs on one
microcontroller~\cite{david2021tflmicro,banbury2021mlperf}. A compact
artificial neural network, such as DS-CNN~\cite{zhang2018helloedge},
BC-ResNet~\cite{kim2021bcresnet}, or the speaker network of Balian et
al.~\cite{balian2021smallfootprint}, computes the embedding from frame-level
features such as mel-frequency cepstral coefficients (MFCCs), and the same core stores the gallery, scores, and
enrolls.

\textbf{Spiking neural networks and neuromorphic processors.}
Neuromorphic audio systems encode sound as sparse events and process them with
a spiking neural network (SNN) of leaky integrate-and-fire (LIF)
neurons~\cite{gerstner2002spiking}. Each neuron's synaptic current and
membrane voltage decay with time constants $\tau_{\mathrm{syn}}$ and
$\tau_{\mathrm{mem}}$, and the neuron spikes when its voltage crosses a
threshold. These decays are a feed-forward SNN's only memory of past input, so
they set how far back each neuron integrates acoustic evidence. Surrogate
gradients carry training through the non-differentiable
spike~\cite{neftci2019surrogate}.
Neuromorphic processors~\cite{akopyan2015truenorth,davies2018loihi,muir2023xylo}
store each weight beside its neuron and compute only when spikes
arrive~\cite{roy2019spikebased}, so the silicon fixes the size and arithmetic
of the network. In Xylo Audio~3~\cite{synsense2023xyloaudio3ds}, an audio front
end converts the microphone signal into a $16$-channel spike raster for its
spiking core. Its limits are typical of the class: at
most $1{,}000$ hidden neurons and $31{,}744$ int8 recurrent synapses; integer
thresholds and $16$-bit saturating state; power-of-two decays; at most $31$
events per hidden neuron and one per readout neuron per
timestep; $32$ readout neurons as the only off-chip output; and one timestep
of delay per hidden-to-hidden hop.

\textbf{Fitting a trained network to neuromorphic hardware.}
Deploying an SNN trained in floating point on a neuromorphic chip involves
three steps: pruning, quantization-aware training, and mapping.
Pruning removes synapses until the network meets the synapse budget; at a
fixed number of weights, pruning a large dense network yields higher accuracy
than training a small dense one~\cite{zhu2017prune}, and
SpikeFit~\cite{spikefit2025} adapts compression-aware fine-tuning to SNNs,
pruning the channels with the lowest Fisher spike contribution.
Quantization-aware training (QAT)~\cite{jacob2018quantization} simulates
the chip's integer weights and bit-shift decays in the forward pass and passes
gradients through the rounding with the straight-through
estimator~\cite{bengio2013ste}. Mapping converts the trained
parameters into the chip configuration, and a cycle-accurate simulator, such
as XyloSim in Rockpool~\cite{muir2023xylo}, predicts the spikes the chip will
produce.

%% file: Sections/3-system_overview_new.tex
\section{System Overview}
\label{sec:system-overview}

\begin{figure*}[t]
  \centering
  \includegraphics[width=0.95\textwidth]{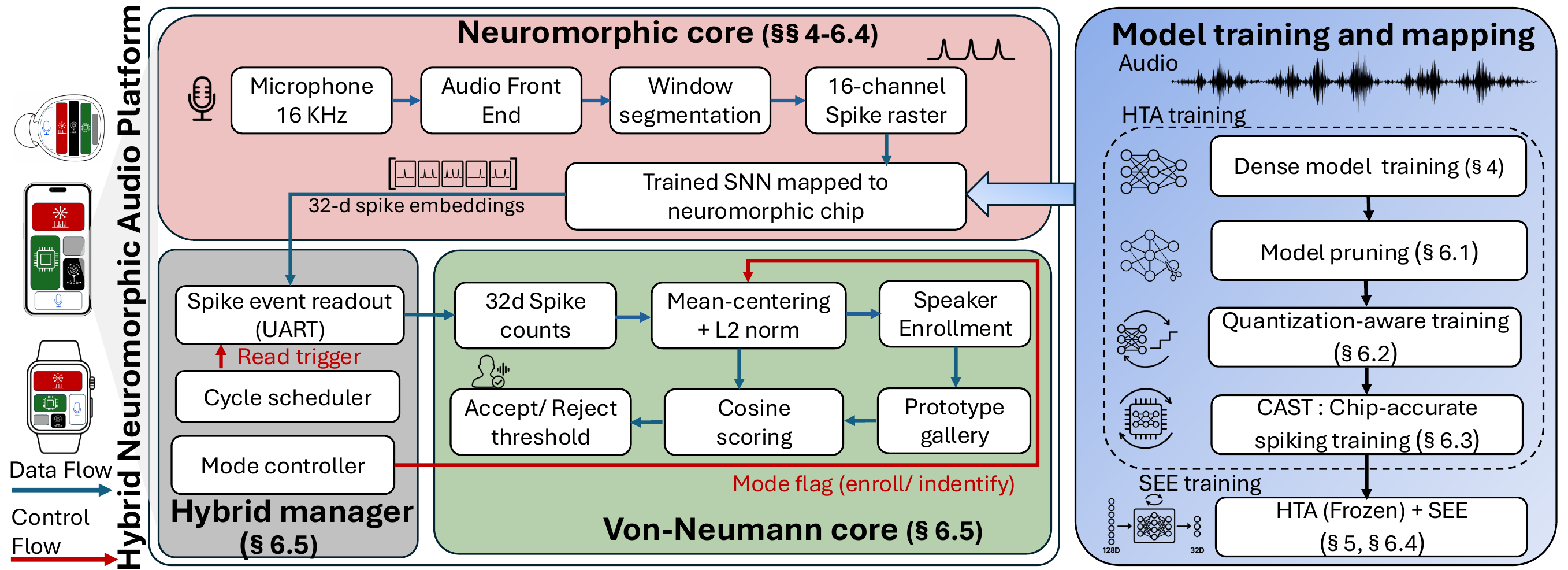}
  \caption{\name system overview: offline training and mapping (right) and
  the inference pipeline (left).}
  \label{fig:hyvox-system-overview}
\end{figure*}

\name splits open-set speaker identification at the embedding. A spiking
network on Xylo Audio~3 embeds each three-second window of audio, and an
STM32L476 host keeps every operation that needs general-purpose compute.
Fig.~\ref{fig:hyvox-system-overview} shows the two parts of the system: an
offline pipeline that trains the spiking network and fits it to the chip
(right), and the hybrid neuromorphic audio platform that runs it on the device
(left). Our four contributions form the offline pipeline. The HTA-SNN
backbone and the SEE define the network, and dense-to-sparse compression and
CAST fit it to the chip. The neuromorphic--von Neumann platform then runs the
network, with the open-set gallery on the host.

\subsection{Model Training and Mapping}

\textbf{HTA-SNN backbone.}
The backbone is a hierarchical temporal abstraction SNN of $832$ LIF neurons
that learns a synaptic time constant per neuron and carries fast acoustic
evidence to its slow stages through cross-scale projections
(Sec.~\ref{sec:hta-architecture}). It is pretrained as a speaker classifier
on VoxCeleb2 from the $16$-channel spike rasters of the chip's audio front end (AFE), so the network learns from the same input it receives on the device.

\textbf{Dense-to-sparse compression.}
The backbone is trained dense with $188.4$k synapses, $5.9\times$ the chip's
budget. Our per-synapse extension of SpikeFit~\cite{spikefit2025} then
prunes it iteratively to $29.0$k synapses with compression-aware
fine-tuning (Sec.~\ref{sec:compression}), and quantization-aware training replaces the
floating-point LIF update with the chip's int8 weights and bit-shift decays
(Sec.~\ref{sec:bitshift-qat}).

\textbf{Chip-accurate spiking training (CAST).}
The quantized backbone still runs without the chip's event caps and per-hop
delay. CAST fine-tunes it through a PyTorch replica of the chip's integer
neuron update, derived from the cycle-accurate simulator XyloSim, whose
trainable parameters are the chip's own int8 weights and integer thresholds.
Training therefore optimizes the exact configuration that is written
to the chip (Sec.~\ref{sec:cast}).

\textbf{Spiking embedding extractor (SEE).}
With the backbone frozen, the SEE, a $32$-neuron spiking readout, is
pretrained episodically on VoxCeleb1, whose speakers are disjoint from the
backbone's VoxCeleb2 speakers. Each episode
enrolls new speakers from a few windows, identifies further windows of them,
and rejects speakers who are not enrolled, so the readout learns the
operations the device later performs on its users (Sec.~\ref{sec:see}). The
SEE is finally fine-tuned on the chip-exact spikes of the CAST backbone.
The backbone and the SEE are mapped once to a single Xylo configuration that
serves every device; no weight is ever trained on a user's audio.

\subsection{Hybrid Neuromorphic Audio Platform}

\textbf{Neuromorphic core.}
A $16\,\mathrm{kHz}$ microphone feeds the AFE of Xylo Audio~3,
which converts the waveform into a $16$-channel spike raster. Each
three-second window of this raster drives the mapped HTA-SNN and SEE from a
reset state, and the chip emits the spike counts of the $32$ SEE neurons. This
count vector is the only output of the chip and the entire interface to the
host.

\textbf{Hybrid manager.}
The firmware's hybrid manager mediates the chip--host boundary. A cycle
scheduler issues a read trigger at the end of each window, the readout
transfers the $32$ counts over UART, and a mode controller sets the host to
enrollment or identification. One
decision moves $70$ bytes per three-second window, independent of the size of
the network on the chip.

\textbf{von Neumann core.}
The STM32 runs the metric kernel and holds every speaker-specific quantity.
It L2-normalizes each count vector into the trained embedding, averages the
latest three windows, scores the result against every prototype in the
gallery by cosine similarity, and accepts the best match only if its score
reaches a rejection threshold chosen for the current gallery size.
In enrollment mode, the host accumulates the normalized embeddings of a new
speaker's windows into one prototype and writes it to the gallery. Adding or deleting a speaker is a
write to host memory with $64$ bytes per speaker, and the network on the chip
is unchanged, so each household differs from another only in the gallery on
its host (Sec.~\ref{sec:host-inference}).

%% file: Sections/4-technique1.tex
\section{Hierarchical Temporal Abstraction (HTA)}
\label{sec:hta-architecture}

Speaker identity includes multiple temporal scales: short-timescale acoustic events (plosive releases, formant transitions, fricative onsets) cue vocal-tract structure, while longer patterns (speaking rate, pitch contour, slowly varying resonance) give more stable evidence over hundreds of milliseconds. Modern encoders capture this range with deep, over-parameterised temporal models~\cite{snyder2018xvectors,desplanques2020ecapa}, but current neuromorphic hardware, including Xylo Audio~3~\cite{synsense2023xyloaudio3ds}, caps the deployed network at $1{,}000$ neurons, $31{,}744$ recurrent synapses, and low-precision integer weights. Naively reducing conventional architectures to this regime strips precisely the temporal capacity needed for robust discrimination; the architectural problem is thus not just fewer neurons, but allocating the remaining state to the right timescales. To overcome this, we propose \emph{hierarchical temporal abstraction}
(HTA), a compact LIF hierarchy for speaker embeddings. Its layers learn
distinct integration windows, allowing fast acoustic cues and slower speaker
patterns to be represented in the same model. Cross-scale projections carry
short-timescale evidence forward while deeper layers accumulate long-range
speaker context.

\subsection{HTA's Concept}

Our key observation is that the temporal depth in a spiking network is
not determined only by the number of layers, but also by the temporal filtering
applied before each neuron integrates its input. Hence, we distinguish the two LIF
timescales explicitly. The membrane time constant $\tau_{\mathrm{mem}}$
controls how long a neuron retains voltage after receiving current; in our
model it is fixed at $2$ timesteps for every neuron. The synaptic time constant $\tau_{\mathrm{syn}}$ controls
the temporal content delivered to the membrane by filtering presynaptic spike
trains. This is the trainable quantity in HTA. Small
$\tau_{\mathrm{syn}}$ values preserve sharp transients, while larger
$\tau_{\mathrm{syn}}$ values expose the neuron to evidence integrated over
longer intervals~\cite{gerstner2002spiking}. Rather than assigning a single
fixed input timescale to the entire network, HTA distributes trainable
synaptic filters across a sequential hierarchy.

\begin{figure}[h]
  \centering
  \includegraphics[width=0.85\linewidth]{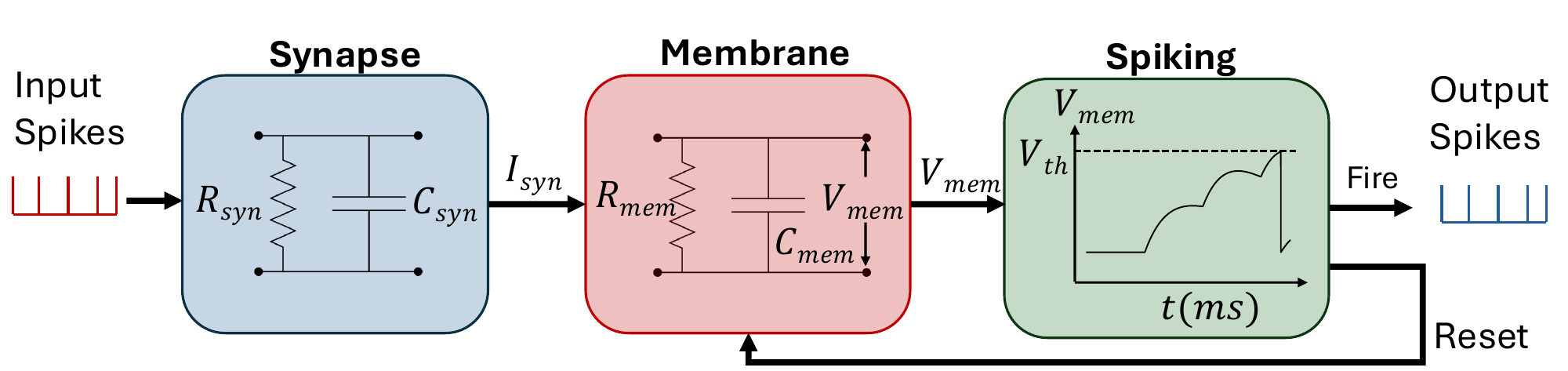}
  \caption{LIF dynamics with trainable synaptic decay and fixed membrane
  decay.}
  \label{fig:lif-dynamics}
\end{figure}

Each LIF layer computes
\begin{align}
    \mathbf{I}_{\mathrm{syn}}^{(\ell)}(t) &=
        \boldsymbol{\beta}_{\mathrm{syn}}^{(\ell)} \odot \mathbf{I}_{\mathrm{syn}}^{(\ell)}(t-1)
        + \mathbf{W}^{(\ell)} \mathbf{s}^{(\ell-1)}(t), \\
    \mathbf{V}_{\mathrm{mem}}^{(\ell)}(t) &=
        \alpha_{\mathrm{mem}} \mathbf{V}_{\mathrm{mem}}^{(\ell)}(t-1)
        + \mathbf{I}_{\mathrm{syn}}^{(\ell)}(t), \\
    \mathbf{s}^{(\ell)}(t) &= H\!\left(
        \mathbf{V}_{\mathrm{mem}}^{(\ell)}(t) - \vartheta
    \right),
\end{align}
with a subtractive reset after threshold crossing. Here
$\ell$ is the layer, $t$ is time, $\mathbf{s}^{(\ell)}(t)$ is the
output spike vector, $\mathbf{I}_{\mathrm{syn}}^{(\ell)}(t)$ is the filtered
synaptic current, $\mathbf{V}_{\mathrm{mem}}^{(\ell)}(t)$ is the membrane voltage,
$\mathbf{W}^{(\ell)}$ is the feed-forward weight matrix, $\vartheta$ is
the firing threshold, $H(\cdot)$ is the Heaviside step function, and $\odot$
is the element-wise product.
The vector $\boldsymbol{\beta}_{\mathrm{syn}}^{(\ell)}$ holds one synaptic
decay per neuron and is set element-wise by
$\boldsymbol{\tau}_{\mathrm{syn}}^{(\ell)}\in\mathbb{R}^{N_\ell}$, where
$N_\ell$ is the width of layer $\ell$, whereas the membrane decay
$\alpha_{\mathrm{mem}}$ is determined by the fixed $\tau_{\mathrm{mem}}$:
\begin{equation}
    \boldsymbol{\beta}_{\mathrm{syn}}^{(\ell)}
    = \exp\!\left(-\frac{\Delta t}{\boldsymbol{\tau}_{\mathrm{syn}}^{(\ell)}}\right),
    \qquad
    \alpha_{\mathrm{mem}}
    = \exp\!\left(-\frac{\Delta t}{\tau_{\mathrm{mem}}}\right).
\end{equation}
We learn
$\boldsymbol{\tau}_{\mathrm{syn}}^{(\ell)}$, one value per neuron, in the four
stages after the input projection, using surrogate-gradient
backpropagation~\cite{neftci2019surrogate}; the $64$ neurons of the input
projection keep a fixed $\tau_{\mathrm{syn}}=2$. Thus HTA learns what temporal
content each layer receives, while keeping voltage retention fixed for
hardware-stable LIF dynamics. Fig.~\ref{fig:lif-dynamics} illustrates this
three-stage LIF flow of synapse, membrane, and spiking. Incoming spikes are first shaped by the trainable synaptic
decay $\boldsymbol{\beta}_{\mathrm{syn}}^{(\ell)}$, producing
$\mathbf{I}_{\mathrm{syn}}^{(\ell)}(t)$. The resulting current is then
integrated by membrane dynamics with fixed decay $\alpha_{\mathrm{mem}}$.
Finally, when $\mathbf{V}_{\mathrm{mem}}^{(\ell)}(t)$ crosses the threshold
$\vartheta$, the neuron emits $\mathbf{s}^{(\ell)}(t)$ and resets its
membrane state.

This LIF primitive is then repeated with different learned
$\tau_{\mathrm{syn}}$ values to instantiate HTA as the proposed
speaker-embedding backbone. The backbone is
deliberately sequential: parallel multi-branch designs can cover multiple
timescales, but they duplicate neurons and synapses --- poorly matched
to the Xylo resource envelope. Our backbone instead reuses a single stream of
spiking state: early layers form fast local descriptors, intermediate layers
integrate over broader acoustic patterns, and deeper layers accumulate
speaker-level context. Cross-scale skip projections then expose early detail
to later integrators without introducing a full parallel pathway.

\subsection{HTA's Architecture}

\subsubsection{Proposed Backbone}
\label{sec:backbone}
\begin{figure}[t]
  \centering
  \includegraphics[width=0.8\linewidth]{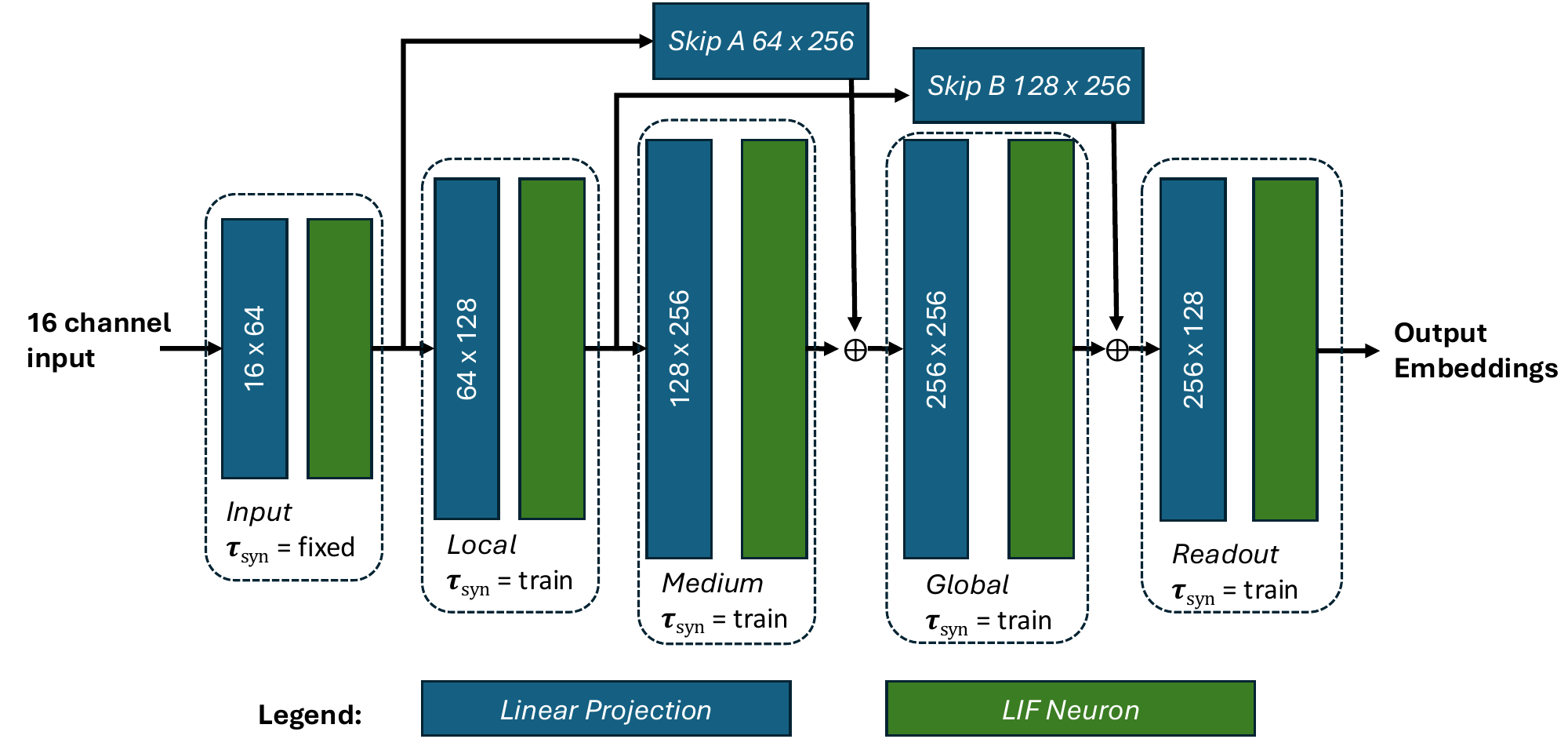}
  \caption{HTA backbone with sequential multi-timescale LIF stages and
  projected cross-scale fusion.}
  \label{fig:hta-backbone}
\end{figure}

Our proposed HTA backbone receives a spike raster
$\mathbf{x} \in \{0,1\}^{T \times 16}$, where the 16 channels are produced by
the Xylo Audio~3 audio front end (AFE) and $T = 293$ timesteps correspond to a
3-second utterance segment. Fig.~\ref{fig:hta-backbone} shows the resulting
single-stream hierarchy, including the projected skip paths that inject early
timescale information into later integrators.

\noindent Algebraically, the network uses five LIF stages:
\begin{align}
    \mathbf{h}_1 &= \operatorname{LIF}_{64}^{\tau_{\mathrm{syn}}=2.0}
        \!\left(\mathbf{W}_{\mathrm{in}}\mathbf{x}\right)
        &&\text{input projection}, \\
    \mathbf{h}_2 &= \operatorname{LIF}_{128}^{\tau_{\mathrm{syn},\mathrm{local}}}
        \!\left(\mathbf{W}_{\mathrm{local}}\mathbf{h}_1\right)
        &&\text{fast local features}, \\
    \mathbf{h}_3 &= \operatorname{LIF}_{256}^{\tau_{\mathrm{syn},\mathrm{med}}}
        \!\left(\mathbf{W}_{\mathrm{med}}\mathbf{h}_2\right)
        &&\text{med-range integration}, \\
    \mathbf{h}_4 &= \operatorname{LIF}_{256}^{\tau_{\mathrm{syn,global}}}
        \!\left(\mathbf{W}_{global}(\mathbf{h}_3 + \mathbf{S}_{A}\mathbf{h}_1)\right)
        &&\text{long-range integration}, \\
    \mathbf{h}_5 &= \operatorname{LIF}_{128}^{\tau_{\mathrm{syn,readout}}}
        \!\left(\mathbf{W}_{readout}(\mathbf{h}_4 + \mathbf{S}_{B}\mathbf{h}_2)\right)
        &&\text{readout integration}.
\end{align}
Here $\mathbf{S}_{A}: 64 \rightarrow 256$ and
$\mathbf{S}_{B}: 128 \rightarrow 256$ are learned linear projections used for
cross-scale fusion. The first projects the input representation into the
256-dimensional long-range stage, while the second projects local features
onto the 256-dimensional pre-readout bus before the $256 \rightarrow 128$
projection $\mathbf{W}_{readout}$. The trainable synaptic time constants
$\boldsymbol{\tau_{\mathrm{syn},\mathrm{local}}}$, $\boldsymbol{\tau_{\mathrm{syn},\mathrm{med}}}$,
$\boldsymbol{\tau_{\mathrm{syn,global}}}$, and $\boldsymbol{\tau_{\mathrm{syn,readout}}}$ are
initialized log-uniformly and constrained to a fixed admissible range during
training. Before the last compression step, they are snapped to the chip's
power-of-two grid and frozen (Sec.~\ref{sec:compression}).

The backbone embedding is the L2-normalized mean spike rate of the final layer:
\begin{equation}
\small 
    \bar{\mathbf{h}}_5 = \frac{1}{T}\sum_{t=1}^{T}\mathbf{h}_{5}(t),
    \qquad
    \mathbf{z} =
    \frac{\bar{\mathbf{h}}_5}{\|\bar{\mathbf{h}}_5\|_2}.
\end{equation}
Here $\mathbf{h}_{5}(t)$ denotes the 128-dimensional spike vector emitted by
the final LIF stage at timestep $t$, and $\mathbf{z}$ is the unit-norm
backbone embedding. It is the output with which the backbone is pretrained and
measured (Sec.~\ref{sec:eval-method}); the SEE turns the spikes of
$\mathbf{h}_5$ into the deployed speaker embedding $\mathbf{e}$
(Sec.~\ref{sec:see}). Rate pooling suits the chip, since Xylo exposes spike
counts directly and pooling turns a spike train of any duration into a
fixed-dimensional vector.

\subsubsection{Cross-Scale Fusion}

A strictly feed-forward temporal hierarchy creates a failure mode {\em i.e.} fast
phonetic evidence must pass through progressively slower layers before it can
affect the embedding. By the time it reaches the readout, transient structure
may have been attenuated by repeated integration. HTA mitigates this bottleneck
with two projected skip connections. The first injects the input projection
$\mathbf{h}_1$ into the long-range integrator, giving the global layer access
to unsmoothed spectral evidence. The second injects the local representation
$\mathbf{h}_2$ into the readout integrator, preserving short-timescale
articulatory cues alongside deeper context vector.

These skips are not residual blocks in the conventional deep-network sense,
they connect different channel widths and different temporal regimes. Their
role is cross-scale fusion under a fixed neuron budget. The final embedding
therefore reflects three complementary sources of information {\em i.e.} fast local
events from early layers, intermediate acoustic trajectories, and long-range
speaker context from the deeper integrators.

\subsubsection{Hardware Compliance}

The backbone contains
$64 + 128 + 256 + 256 + 128 = 832$ LIF neurons, within Xylo's $1{,}000$ hidden
neurons, and the SEE (Sec.~\ref{sec:see}) occupies the chip's $32$ readout
neurons. The final 128-neuron
backbone layer also respects the constraint that the layer driving the
hardware readout interface must not exceed 128 neurons. The architecture
therefore fits the chip in its neuron count, its readout width, and its
integer spike-rate pooling, and the embedding comes from the $32$ readout
neurons, the only neurons whose spikes leave the chip during deployment.

\subsubsection{Weight Initialisation}

Trainable synaptic time constants couple the scale of the weights to the
timescale of each neuron. Under the synaptic update above, a presynaptic spike
of weight $w$ adds a current that decays as $w\beta_{\mathrm{syn}}^{k}$ over
the following timesteps $k$, so it delivers a total charge of
$w/(1-\beta_{\mathrm{syn}}) \approx w\,\tau_{\mathrm{syn}}/\Delta t$ to the
membrane. The initial time constants span $2$ to $256$ timesteps, so with the
same random initialization in every layer the slowest neurons would receive up
to $128\times$ the drive of the fastest, saturating while fast neurons barely
fire. Following SynNet~\cite{muir2023xylo,bos2024micropower}, the reference SNN for Xylo, we
divide out this gain at initialization,
\begin{equation}
\small
    W^{(\ell)}_{0,ij} =
    \frac{\Delta t}{\tau^{(\ell)}_{\mathrm{syn},0,i}}\,\tilde{W}^{(\ell)}_{ij},
\end{equation}
where $i$ indexes the neuron that the weight drives and $j$ the presynaptic
neuron, $\tilde{W}^{(\ell)}_{ij}$ is the random initialization,
$\tau^{(\ell)}_{\mathrm{syn},0,i}$ is the initial synaptic time constant of
neuron $i$, and $\Delta t$ is one timestep, the unit in which all time constants are expressed. Every neuron
then receives about the same total charge per input spike, so neurons of every
timescale take part in learning from the first training step. The backbone is pretrained on VoxCeleb2 with the additive angular margin softmax (AAM-Softmax) loss~\cite{deng2019arcface} and a
regularizer on $\log\boldsymbol{\tau}_{\mathrm{syn}}$ that keeps the mean
timescales of the local, medium, and global stages ordered with minimum gaps
and keeps a minimum spread of timescales within each stage. Training uses Adam
with warmup followed by cosine learning-rate decay.

%% file: Sections/5-technique2_SEE.tex
\section{Spiking Embedding Extractor (SEE)}
\label{sec:see}
\label{sec:metric-learning}

Pretrained on VoxCeleb2 (Sec.~\ref{sec:hta-architecture}), the HTA backbone
has learned to extract features that distinguish one speaker from another.
However, it learns them as a closed-set classifier, separating a fixed set of
training speakers, whereas the deployed system faces an open-set problem. Its
users are never part of training: a user is enrolled on the device from a short
enrollment recording (Sec.~\ref{sec:see-gallery}), and no network weight is
updated with that audio. In operation the device also hears guests who were never enrolled. The
system must therefore recognise an enrolled speaker from a representation built
from that recording and turn away audio from anyone else. Closed-set training
exercises neither requirement: the classifier never builds a speaker
representation from a few examples, and never encounters speakers outside its
training set, whose voices may fall anywhere between the learned classes. Open-set recognition thus calls for an embedding
extractor that has learned, from other speakers, how to separate speakers given
only a few shots each, and how to separate them from speakers who are not
enrolled. To this end, we propose the \emph{spiking embedding extractor} (SEE),
a 32-neuron spiking readout on top of the frozen backbone. The SEE is trained
episodically on a disjoint speaker corpus: each episode enrolls a new set of
training speakers from a few windows, identifies further windows of those
speakers, and rejects speakers who are not enrolled, the same operations the
deployed system later performs on its users without any further training.

\subsection{SEE's Concept}
\label{sec:see-concept}

Our key observation is that, once the backbone is fixed, recognizing a speaker
on the device takes three steps in one embedding space: average a speaker's
enrollment embeddings into a prototype, find the prototype closest to a query by
cosine similarity, and reject the query if that similarity is below a threshold.
The SEE is trained to produce embeddings for exactly these steps, so the network
used at inference is the network trained, with no classifier or scoring layer added.

Let $f_{\theta}$ denote the frozen HTA backbone. For an AFE spike raster
$\mathbf{x}\in\mathbb{Z}^{T\times 16}$, the backbone emits the spike sequence
$f_{\theta}(\mathbf{x})=[\mathbf{z}_1,\ldots,\mathbf{z}_T]$ with
$\mathbf{z}_t\in\{0,1\}^{128}$. The SEE is a bias-free linear projection
followed by a 32-neuron LIF layer:
\begin{align}
\small
    \mathbf{u}_t &= \mathbf{W}_{s}\mathbf{z}_t, \\
    \mathbf{r}_t &= \operatorname{LIF}_{32}^{\tau_{\mathrm{syn}}=2\Delta t}
        (\mathbf{u}_t), \\
    \bar{\mathbf{r}} &=
        \frac{\sum_{t=1}^{T}\mathbf{r}_t}
             {\left\|\sum_{t=1}^{T}\mathbf{r}_t\right\|_2}, \\
    \mathbf{e} &= E(\mathbf{x}) =
        \frac{\bar{\mathbf{r}}-\boldsymbol{\mu}}
             {\left\|\bar{\mathbf{r}}-\boldsymbol{\mu}\right\|_2},
\end{align}
where $\mathbf{W}_{s}\in\mathbb{R}^{32\times 128}$ is the trainable projection,
$\mathbf{u}_t$ is the input current to the readout layer,
$\mathbf{r}_t\in\{0,1\}^{32}$ is the readout spike vector, $\Delta t$ is one
binned AFE timestep, $\bar{\mathbf{r}}$ is the normalised spike count,
$\boldsymbol{\mu}\in\mathbb{R}^{32}$ is a fixed centring vector, and
$\mathbf{e}\in\mathbb{R}^{32}$ is the unit-norm speaker
embedding. We write $E(\cdot)$ for the full extractor, {\em i.e.}, the backbone and the SEE
together. The normalised spike count is identical to
the normalised mean spike rate and to the integer count vector read from Xylo
(Sec.~\ref{sec:host-inference}). Spike counts are non-negative, so every
$\bar{\mathbf{r}}$ lies in the positive orthant and the windows of any two
speakers share a large common component. Subtracting $\boldsymbol{\mu}$, the
mean of $\bar{\mathbf{r}}$ over training windows, removes this component, so
cosine similarities reflect what distinguishes one speaker from another. The
host stores $\boldsymbol{\mu}$ and applies it with 32 subtractions per
window. $\mathbf{W}_{s}$ is learned as an int8
matrix with one trainable scale per output neuron, giving 4,128 trainable
parameters in total. The synaptic time constant $\tau_{\mathrm{syn}}=2\Delta t$
is fixed, so its decay is realised as a single bit shift on the chip.

This design keeps the learning problem deliberately small. The backbone carries
the speaker-discriminative structure learned from VoxCeleb2, and the SEE learns
only how to expose that structure in 32 dimensions so that the same-speaker
windows fall close to their prototype, different speakers fall apart, and
unenrolled speakers stay away from every prototype. Because the SEE never
encodes which speakers are enrolled, the identities it serves can change after
training without changing the network.

\subsection{Pretraining SEE}
\label{sec:see-pretraining}

We pretrain the SEE for two properties: generalization to speakers that were
never seen in training, and rejection of speakers that are not enrolled. Both
follow from few-shot open-set learning (FSOSL)~\cite{kc2022openfeat,
liu2020peeler}, in which the model is trained on episodes that each simulate one
enrollment scenario.

In each episode, we select $N$
speakers with $K$ support windows and $Q$ query windows per speaker, an $N$-way
$K$-shot problem. We denote the support set as
$\mathcal{S}_{\mathrm{seen}}=\{(\mathbf{x}^{S}_i, y^{S}_i)\}_{i=1}^{NK}$ and the
seen query set as
$\mathcal{Q}_{\mathrm{seen}}=\{(\mathbf{x}^{Q}_i, y^{Q}_i)\}_{i=1}^{NQ}$, with
$y_i\in\mathcal{Y}_{S}$. In addition, we select $R$ unseen speakers with $T_u$
windows each, forming the unseen query set
$\mathcal{Q}_{\mathrm{unseen}}=\{\mathbf{x}^{U}_i\}_{i=1}^{RT_u}$ with labels in
$\mathcal{Y}_{U}$, where $\mathcal{Y}_{S}\cap\mathcal{Y}_{U}=\emptyset$. The
support windows play the role of enrollment utterances, and the prototype of
speaker $k$ is the normalised mean of its support embeddings,
\begin{equation}
\small
    \mathbf{c}_k =
    \frac{\sum_{\mathbf{x}\in\mathcal{S}_k} E(\mathbf{x})}
         {\left\|\sum_{\mathbf{x}\in\mathcal{S}_k} E(\mathbf{x})\right\|_2},
\end{equation}
where $\mathcal{S}_k\subset\mathcal{S}_{\mathrm{seen}}$ holds the support
windows of speaker $k$. The unseen speakers play the role of guests: they are
present in the episode but have no prototype. Both groups are resampled in every
episode, so the SEE never sees the same gallery twice and cannot memorise a
particular set of speakers or a fixed background population.

For windows from seen speakers, same-speaker embeddings should be compact and
different speakers well separated. We apply the supervised contrastive
loss~\cite{khosla2020supcon} over
$\mathcal{B}=\mathcal{S}_{\mathrm{seen}}\cup\mathcal{Q}_{\mathrm{seen}}$:
\begin{equation}
\small
    \mathcal{L}_{\mathrm{id}}
    =
    \sum_{i\in\mathcal{B}}
    \frac{-1}{|\mathcal{P}(i)|}
    \sum_{p\in\mathcal{P}(i)}
    \log
    \frac{
        \exp(\mathbf{e}_i^\top\mathbf{e}_p / \tau_c)
    }{
        \sum_{a\in\mathcal{B}\setminus\{i\}}\exp(\mathbf{e}_i^\top\mathbf{e}_a / \tau_c)
    },
\end{equation}
where $\mathcal{P}(i)$ is the set of same-speaker windows of anchor $i$ and
$\tau_c$ is a temperature. Because embeddings are unit-norm, every inner
product is a cosine similarity, the same score used at inference. For windows
from unseen speakers, the model should
not assign a high similarity to any seen speaker. The deployed system rejects a
query whose highest prototype similarity falls below a threshold
(Sec.~\ref{sec:see-gallery}), so we penalise this similarity with a hinge loss:
\begin{equation}
\small
    \mathcal{L}_{\mathrm{rej}}
    =
    \frac{1}{RT_u}
    \sum_{\mathbf{x}\in\mathcal{Q}_{\mathrm{unseen}}}
    \max\!\Bigl(0,\;\max_{k}\,E(\mathbf{x})^\top\mathbf{c}_k - m\Bigr),
\end{equation}
where $m$ is a margin. The total loss is
$\mathcal{L}_{\mathrm{SEE}}=\mathcal{L}_{\mathrm{id}}+\lambda\,\mathcal{L}_{\mathrm{rej}}$,
where $\lambda$ controls the weight of the rejection loss. Both losses act on
centred embeddings. During pretraining, $\boldsymbol{\mu}$ is a running average
of $\bar{\mathbf{r}}$ over the training episodes and receives no gradient, and
its final value is stored with the SEE.

We pretrain on VoxCeleb1~\cite{nagrani2017voxceleb}, whose 1,251 speakers are
disjoint from the VoxCeleb2 speakers used for the backbone, so the SEE learns
from speakers that are also unseen to the backbone, matching the deployment
condition. Speakers are split into 900 training, 200 validation, and 151 test
speakers. Each speaker contributes 60
three-second windows drawn across its videos, converted by the Xylo Audio~3 AFE
simulator into $293\times 16$ spike rasters. In each episode, we set $N=20$,
close to the largest household size we target, with $K=4$, $Q=5$, $R=5$, and
$T_u=5$. We train for 16,000 episodes, averaging gradients over four episodes per
update, with Adam, a learning
rate of 0.03 with cosine decay, and a weight decay of $10^{-3}$, using
$\tau_c=0.1$, $\lambda=1$, $m=0.4$, and an exponential-moving-average momentum of
0.99 for $\boldsymbol{\mu}$. The checkpoint is selected on the validation speakers by
identification accuracy and open-set equal error rate.

\subsection{SEE's Speaker Gallery}
\label{sec:see-gallery}

After pretraining, the SEE is quantised with the backbone and mapped to Xylo,
and its weights do not change again. The identities the system recognises are
held entirely in a speaker gallery on the MCU host,
$\mathcal{G}=\{(k,\mathbf{c}_k)\}_{k=1}^{n}$, with one unit-norm prototype per
enrolled speaker and $n\le N$. The gallery is the only speaker-specific state in
the system.

A speaker is enrolled from 60\,s of speech, sliced into $|\mathcal{E}_k|=29$
three-second windows with a 2\,s stride. Its prototype is the normalised mean of
the window embeddings,
\begin{equation}
\small
    \mathbf{c}_k =
    \frac{\frac{1}{|\mathcal{E}_k|}\sum_{j\in\mathcal{E}_k}\mathbf{e}_j}
         {\left\|\frac{1}{|\mathcal{E}_k|}\sum_{j\in\mathcal{E}_k}\mathbf{e}_j\right\|_2},
\end{equation}
the same operation that builds $\mathbf{c}_k$ from the support set during
pretraining, applied to more windows.

A single three-second query window gives only a noisy estimate of the
speaker. The host therefore forms each identification decision from $W$
consecutive
windows~\cite{xie2019utterance} by averaging their embeddings:
\begin{equation}
\small
    \bar{\mathbf{e}}_q =
    \frac{\frac{1}{W}\sum_{w=1}^{W}\mathbf{e}^{(w)}_q}
         {\left\|\frac{1}{W}\sum_{w=1}^{W}\mathbf{e}^{(w)}_q\right\|_2},
\end{equation}
where $\mathbf{e}^{(w)}_q$ is the embedding of the $w$-th window of the query.
We use $W=3$ over a sliding window: after the first 9\,s, each new 3\,s window
yields a decision from the latest three, so a decision is issued every 3\,s.
The query is scored against every prototype in
the gallery,
\begin{equation}
\small
    s_k = \bar{\mathbf{e}}_q^{\top}\mathbf{c}_k,
    \qquad
    \hat{k} = \arg\max_{k\in\mathcal{G}} s_k,
    \qquad
    s_{\max} = s_{\hat{k}},
\end{equation}
and the open-set decision thresholds the best score:
\begin{equation}
\small
    \mathrm{decision}(\mathbf{x}_q) =
    \begin{cases}
        \hat{k}, & s_{\max} \ge \vartheta_{\mathrm{rej}},\\
        \textsc{unknown}, & \text{otherwise},
    \end{cases}
\end{equation}
where $\vartheta_{\mathrm{rej}}$ is the rejection threshold. A query from an
enrolled speaker is correct only if it is both accepted and assigned to the
right prototype; a query from a guest is correct only if it is rejected. The
statistic being thresholded, $s_{\max}$, is the same maximum prototype
similarity that $\mathcal{L}_{\mathrm{rej}}$ suppresses for unseen speakers
during pretraining. Scoring costs one 32-dimensional dot product per enrolled
speaker, and the host kernel that computes it is described in
Sec.~\ref{sec:host-inference}.

\subsection{Runtime Enrollment using SEE}
\label{sec:see-enrollment}
\label{sec:runtime-enrollment}

Runtime enrollment adds a speaker to the gallery without modifying the network.
Because the SEE was pretrained on episodes that each enroll a new set of
speakers, runtime enrollment performs the same operation as every training
episode, applied to a speaker the SEE has never seen. No gradient step, fine-tuning, or reflashing of Xylo is
required.

For a new speaker $k_{\mathrm{new}}$, the device records the enrollment speech,
Xylo embeds each window with the fixed backbone and SEE, and the host computes
$\mathbf{c}_{k_{\mathrm{new}}}$ from these embeddings exactly as in
Sec.~\ref{sec:see-gallery}. The prototype is then appended to the gallery,
\begin{equation}
\small
    \mathcal{G} \leftarrow \mathcal{G} \cup
    \{(k_{\mathrm{new}}, \mathbf{c}_{k_{\mathrm{new}}})\}.
\end{equation}
The prototypes of the existing speakers are not recomputed, so enrollment order
does not matter and adding one speaker never disturbs another. The next query is
scored against the updated gallery, and the new speaker is both identified and
defended against guests by the same threshold as every other member. Removing a
speaker is the inverse operation, deleting its prototype. Enrollment therefore
costs only the forward passes over the enrollment windows and one 32-dimensional
write to host memory; its STM32 implementation is described in
Sec.~\ref{sec:host-inference}.

The pretrained SEE is shared by every deployment. Two devices with different
households run identical networks and differ only in their galleries. The
network is retrained and reflashed only if the backbone or SEE itself is
updated, in which case every stored prototype belongs to the previous embedding
space and all speakers are re-enrolled.

%% file: Sections/6-technique3_new.tex
\section{Chip-Accurate Training for Neuromorphic Deployment}
\label{sec:chip-exact}

The pretrained HTA-SNN and SEE (Sec.~\ref{sec:hta-architecture} and
Sec.~\ref{sec:see}) are floating-point networks. Xylo Audio~3 runs a different
network: $31{,}744$ int8 recurrent synapses, integer thresholds, power-of-two
decays, at most $31$ events per hidden neuron and timestep, one timestep of
delay per hidden-to-hidden hop, and $16$-bit saturating state. Each of these
limits removes speaker information, and the last three are invisible to any
model that does not compute exactly as the chip does. We close this gap in
four steps (Fig.~\ref{fig:hyvox-system-overview}, right). Dense-to-sparse
compression fits the synapse budget (Sec.~\ref{sec:compression}), bitshift
quantization-aware training fits the integer dynamics
(Sec.~\ref{sec:bitshift-qat}), and chip-accurate spiking training (CAST), which
trains through a replica of the chip, fits the event limits, hop delays, and
saturation (Sec.~\ref{sec:cast}). The SEE is then fine-tuned through the CAST
replica's readout on the spikes the chip produces (Sec.~\ref{sec:chip-readout}). The three backbone
steps train on VoxCeleb2, the backbone's pretraining data, and the SEE
fine-tuning trains on the VoxCeleb1 training speakers of the SEE's pretraining
(Sec.~\ref{sec:see-pretraining}). Every step starts from the model of the step
before, so the network that reaches the chip is the network that was trained
for it.

\subsection{Dense-to-Sparse Compression}
\label{sec:compression}

At a fixed number of weights, a large sparse network outperforms a small dense
one~\cite{zhu2017prune}, and the HTA backbone needs its $832$ neurons to cover
the timescales of speaker evidence (Sec.~\ref{sec:hta-architecture}). We
therefore train the backbone dense, with $188.4$k synapses, $5.9\times$ the
chip's budget, and compress it by extending SpikeFit~\cite{spikefit2025}.
SpikeFit removes whole channels, and its authors evaluate it at a single
pruning ratio on an image classifier that runs for $4$
timesteps~\cite{spikefit2025}. In our case, the chip's budget counts individual synapses, and our
backbone runs for $293$ timesteps and must lose $84.6\%$ of its synapses,
so we lift SpikeFit's channel score to a per-synapse score, rank it
globally across stages and skip connections, and prune iteratively.
SpikeFit scores each neuron $c$ by its Fisher spike contribution,
\begin{equation}
\small
    F_c = \frac{1}{N}\sum_{b,t}
    \Bigl(\frac{\partial \mathcal{L}}{\partial y_{b,t,c}}\Bigr)^{2}
    y_{b,t,c}^{2},
\end{equation}
where $y_{b,t,c}$ is the neuron's spike output on calibration window $b$ at
timestep $t$ and $\mathcal{L}$ is the AAM-Softmax pretraining loss. We
extend this score to synapses. A synapse
from neuron $j$ to neuron $i$ receives the score
$|W_{ij}|\,a_j\,F_i$, where $W_{ij}$ is its weight and $a_j$ is the firing
rate of the presynaptic neuron,
so a synapse survives when it is strong, carries spikes, and feeds a neuron
that matters to the loss. Synapses are ranked across all layers and removed in
$14$ steps, each followed by compression-aware fine-tuning. This fine-tuning,
SpikeFit's clusterization-aware training~\cite{spikefit2025}, gives every layer a codebook of $256$ learnable weight values, initialized by
k-means on the layer's weights, so each weight can be stored as an 8-bit index.
In the forward pass every weight is replaced by its nearest codebook value,
gradients reach the full-precision weight through the straight-through
estimator~\cite{bengio2013ste}, and a commitment loss, the mean squared distance
between each weight and its codebook value, moves every codebook value toward
the weights assigned to it. The global ranking lets each stage keep the
synapses it needs, so the network decides where its budget goes. The result
has $29.0$k synapses, $6.5\times$ fewer than the dense backbone.

Before the last pruning step, every learned time constant is snapped to the
chip's bit-shift grid and frozen,
\begin{equation}
\small
    d_{\mathrm{syn}} =
    \operatorname{round}\bigl(\log_2 \tau_{\mathrm{syn}}\bigr),
    \qquad
    \hat{\tau}_{\mathrm{syn}} = 2^{d_{\mathrm{syn}}},
\end{equation}
where $\tau_{\mathrm{syn}}$ is in timesteps and $d_{\mathrm{syn}}$ is the
integer bit shift that the chip stores per neuron, clipped to its range
$1 \le d_{\mathrm{syn}} \le 8$. Each timestep, the chip decays the synaptic
current $I$ by $I \leftarrow I - \lfloor I/2^{d_{\mathrm{syn}}} \rfloor$,
which approximates an exponential decay with time constant
$2^{d_{\mathrm{syn}}}$. The final fine-tuning recovers accuracy under the timescales the chip will
run, and the learned allocation of fast, medium, and slow neurons carries over
to the chip.

\subsection{Bitshift Quantization-Aware Training}
\label{sec:bitshift-qat}

The chip decays both the synaptic current and the membrane voltage with the
integer bit shift $x \leftarrow x - \lfloor x/2^{d}\rfloor$, where
$d \in \{d_{\mathrm{syn}}, d_{\mathrm{mem}}\}$ is the shift of the decaying
state. This update differs from the floating-point update
$x \leftarrow (1-2^{-d})\,x$ by its rounding at every timestep, and these rounding errors accumulate over the $293$ timesteps of a
window. Bitshift quantization-aware training replaces the floating-point LIF
update of every neuron with this integer update, passes gradients through the
rounding with the straight-through estimator~\cite{bengio2013ste}, and
fine-tunes the compressed backbone with the AAM-Softmax objective. The result
is a backbone whose weights, decays, and state updates are all integer.

\subsection{Chip-Accurate Spiking Training (CAST)}
\label{sec:cast}

\textbf{The remaining gap.} After bitshift QAT the backbone uses the chip's
integer arithmetic, but three chip limits are still absent from training: the
hop delays of the chip's recurrent layout, the per-timestep event caps, and
16-bit state saturation. Training updates the backbone as a stack of layers
within one timestep. Xylo runs the hidden neurons as one
recurrent population with a single shared weight matrix, so the skip projections of HTA are folded into direct synapses
(Sec.~\ref{sec:eval-ondevice} gives the resulting synapse count) and
every hidden-to-hidden hop costs one timestep. Skip~A reaches the global stage
in one hop and the main path in three, and likewise for skip~B into the
readout stage, so evidence that training adds at the same timestep arrives on
the chip two timesteps apart. Each hidden neuron emits at most $31$ events per
timestep, each input channel carries at most $15$ events per bin, and the
synaptic current and membrane voltage saturate at $16$ bits. A network trained
without these limits learns to rely on dynamics the chip cannot reproduce, and
mapped directly to the chip, the QAT backbone loses half of its top-1 accuracy
(Sec.~\ref{sec:eval-ondevice}). Imposing the event caps in the training model
alone does not close this gap, since the delays, the folded skips, and the
integer mapping still move every neuron. Training therefore has to see the
chip's computation itself.

\textbf{The CAST replica.} The chip's cycle-accurate simulator, XyloSim, reproduces this
computation but provides no gradients. We propose chip-accurate spiking
training (CAST), which trains the backbone through a differentiable PyTorch
replica of the chip's neuron update, with the update rule derived from
XyloSim. This subsection covers the hidden network, and
Sec.~\ref{sec:chip-readout} the readout. Running XyloSim on the mapped backbone, we record the synaptic
current, membrane voltage, and event count of all $832$ hidden neurons at
every timestep and identify the integer update that reproduces them:
\begin{equation}
\small
\label{eq:cast}
\begin{aligned}
    \mathbf{J}_t &= \mathrm{sat}_{16}\bigl(\mathbf{I}_{t-1}
      + \mathbf{W}_{\mathrm{in}}^{\top}\mathbf{u}_t
      + \mathbf{W}_{\mathrm{rec}}^{\top}\mathbf{S}_{t-1}\bigr),
    \quad
    \mathbf{I}_t = \mathbf{J}_t - \lfloor \mathbf{J}_t / 2^{\mathbf{d}_{\mathrm{syn}}} \rfloor, \\
    \mathbf{U}_t &= \mathrm{sat}_{16}\bigl(\mathbf{V}_{t-1}
      - \lfloor \mathbf{V}_{t-1} / 2^{\mathbf{d}_{\mathrm{mem}}} \rfloor
      + \mathbf{I}_t\bigr), \\
    \mathbf{S}_t &= \mathrm{clip}\bigl(\lfloor \mathbf{U}_t / \boldsymbol{\theta} \rfloor, 0, 31\bigr),
    \quad
    \mathbf{V}_t = \mathbf{U}_t - \mathbf{S}_t \odot \boldsymbol{\theta},
\end{aligned}
\end{equation}
where all quantities are integer vectors over the $832$ hidden neurons.
$\mathbf{I}_t$ is the synaptic current and $\mathbf{V}_t$ the membrane
voltage after timestep $t$; $\mathbf{J}_t$ is the synaptic current after the
new input is added and before it decays, and $\mathbf{U}_t$ is the membrane
voltage before the spike and its reset. $\mathbf{S}_t \in \{0,\ldots,31\}^{832}$
is the number of events each neuron emits, and
$\mathbf{u}_t \in \{0,\ldots,15\}^{16}$ is the capped input raster.
$\mathbf{W}_{\mathrm{in}}$ and $\mathbf{W}_{\mathrm{rec}}$ are the int8 input
and recurrent weights, $\boldsymbol{\theta}$ holds one integer threshold per
neuron, $\mathbf{d}_{\mathrm{syn}}$ and $\mathbf{d}_{\mathrm{mem}}$ are the
per-neuron bit shifts of the two decays.
$\mathrm{clip}(x, a, b) = \min(\max(x, a), b)$ bounds each component to
$[a, b]$, so a neuron emits no events at negative voltage and at most $31$
otherwise, and $\mathrm{sat}_{16}$ clips its argument to the signed $16$-bit
range. The spikes $\mathbf{S}_{t-1}$ of the
previous timestep enter through the shared recurrent matrix, which realizes
the one-timestep hop delay. Driven by the chip's own state, the replica reproduces
$99.8\%$ of XyloSim's per-step hidden spike counts exactly.

\textbf{Training the chip configuration.} The trainable parameters of the replica
are the chip's configuration values: the int8 input and recurrent weights
$\mathbf{W}_{\mathrm{in}}$ and $\mathbf{W}_{\mathrm{rec}}$ and the integer
thresholds $\boldsymbol{\theta}$. They are initialized from the mapped
QAT backbone, held in integer units, and rounded and clipped to the chip's
legal ranges in every forward pass, with straight-through gradients. The
sparsity mask $\mathbf{M}$ of the compressed backbone is fixed and applied to
$\mathbf{W}_{\mathrm{rec}}$ in every forward pass, so a synapse may fall to
zero but no pruned synapse returns, and the synapse budget holds throughout.
The event count is differentiated with a surrogate of slope
$1/\theta_c$ inside the count's range and exponential tails below zero and
above the cap, and each threshold receives its gradient through the
subtractive reset. Per-neuron thresholds act as per-neuron gains, so training
can move strongly driven neurons out of the $31$-event cap and weakly driven
ones into firing. The backbone is fine-tuned with the AAM-Softmax
objective, every forward pass running through the replica
(Algorithm~\ref{alg:cast}). Since the trained parameters are the chip's
configuration, the result maps to the chip with no further quantization. The
method needs only a cycle-accurate simulator, so it applies to any digital
neuromorphic chip that provides one.

\begin{algorithm}[t]
\caption{Chip-accurate spiking training (CAST)}
\label{alg:cast}
\small
\SetKwFor{ForEach}{for each}{do}{end}
\KwIn{chip configuration $\mathbf{W}_{\mathrm{in}}, \mathbf{W}_{\mathrm{rec}}, \boldsymbol{\theta}$
  mapped from the QAT backbone; sparsity mask $\mathbf{M}$;
  training set $\mathcal{D}$ of windows, each an input raster
  $\mathbf{u} = (\mathbf{u}_1, \ldots, \mathbf{u}_T)$ with speaker label $y$;
  AAM-Softmax head $h$}
\KwOut{fine-tuned chip configuration}
\ForEach{\textup{minibatch} $(\mathbf{u}, y) \in \mathcal{D}$}{
  round $\mathbf{W}_{\mathrm{in}}, \mathbf{W}_{\mathrm{rec}}, \boldsymbol{\theta}$ to integers;
  apply mask $\mathbf{M}$ to $\mathbf{W}_{\mathrm{rec}}$\;
  $\mathbf{I}, \mathbf{V}, \mathbf{S} \gets \mathbf{0}$\;
  \For{$t \gets 1$ \KwTo $T$}{
    update $\mathbf{I}, \mathbf{V}, \mathbf{S}$ with Eq.~\eqref{eq:cast}\tcp*{one chip timestep}
  }
  $\mathbf{e} \gets$ L2-normalized mean of $\mathbf{S}_t$ over the last HTA stage\;
  $\mathcal{L} \gets \mathcal{L}_{\mathrm{AAM}}(h(\mathbf{e}), y)$\;
  update $\mathbf{W}_{\mathrm{in}}, \mathbf{W}_{\mathrm{rec}}, \boldsymbol{\theta}, h$
    from $\nabla\mathcal{L}$\tcp*{surrogate gradients}
  clip weights to $[-128, 127]$ and thresholds to $[1, 2^{15}-1]$\;
}
\KwRet{\textup{rounded} $\mathbf{W}_{\mathrm{in}}, \mathbf{W}_{\mathrm{rec}}, \boldsymbol{\theta}$
  \textup{of the best model on held-out speakers}}
\end{algorithm}

\textbf{Relation to hardware-aware training.} Hardware-in-the-loop training
runs the forward pass on the chip and computes gradients through a software
model of it~\cite{schmitt2017hardwareloop,cramer2022surrogate}. It suits analog
chips such as BrainScaleS-2, whose device mismatch no simulator reproduces. A
digital neuromorphic chip computes deterministically, so CAST replaces the chip
with a software replica of its integer update. Training then needs no hardware, runs
at GPU speed on millions of windows, and computes its gradients through the
same arithmetic as its forward pass. The training tools that accompany digital
chips, such as the Rockpool library for Xylo~\cite{muir2023xylo}, model
individual constraints on a layered network, here bit-shift decays and event
caps. The constraints that come from the chip's recurrent layout, the folded
skips, hop delays, shared int8 scales, and integer thresholds, stay outside
training and cost the bitshift-QAT backbone $31.1$ top-1 points on Xylo
Audio~3 (Table~\ref{tab:chip-loss}), which CAST recovers.

\subsection{Fine-Tuning the SEE with CAST}
\label{sec:chip-readout}

The SEE's $32$ readout neurons follow the integer update of
Eq.~\eqref{eq:cast} with two differences: each emits at most one event per
timestep, and its leak removes at least one unit from any positive state. We
extend the CAST replica to the readout with these two rules, and its trainable
parameters are again the chip's int8 weights and integer thresholds. The readout replica reproduces all of XyloSim's
readout spikes. With the CAST backbone fixed, XyloSim produces its
spikes once for every training window, and the SEE is fine-tuned on them
through the readout replica with the episodic objective of its pretraining
(Sec.~\ref{sec:see-pretraining}). The readout thus learns the spikes the chip
actually produces. The backbone and the SEE are written to the chip as one
configuration of int8 weights, integer thresholds, and bit-shift decays,
shared by every device.

\subsection{Hybrid Inference on the Host}
\label{sec:host-inference}

\begin{figure}[t]
  \centering
  \includegraphics[width=0.65\linewidth]{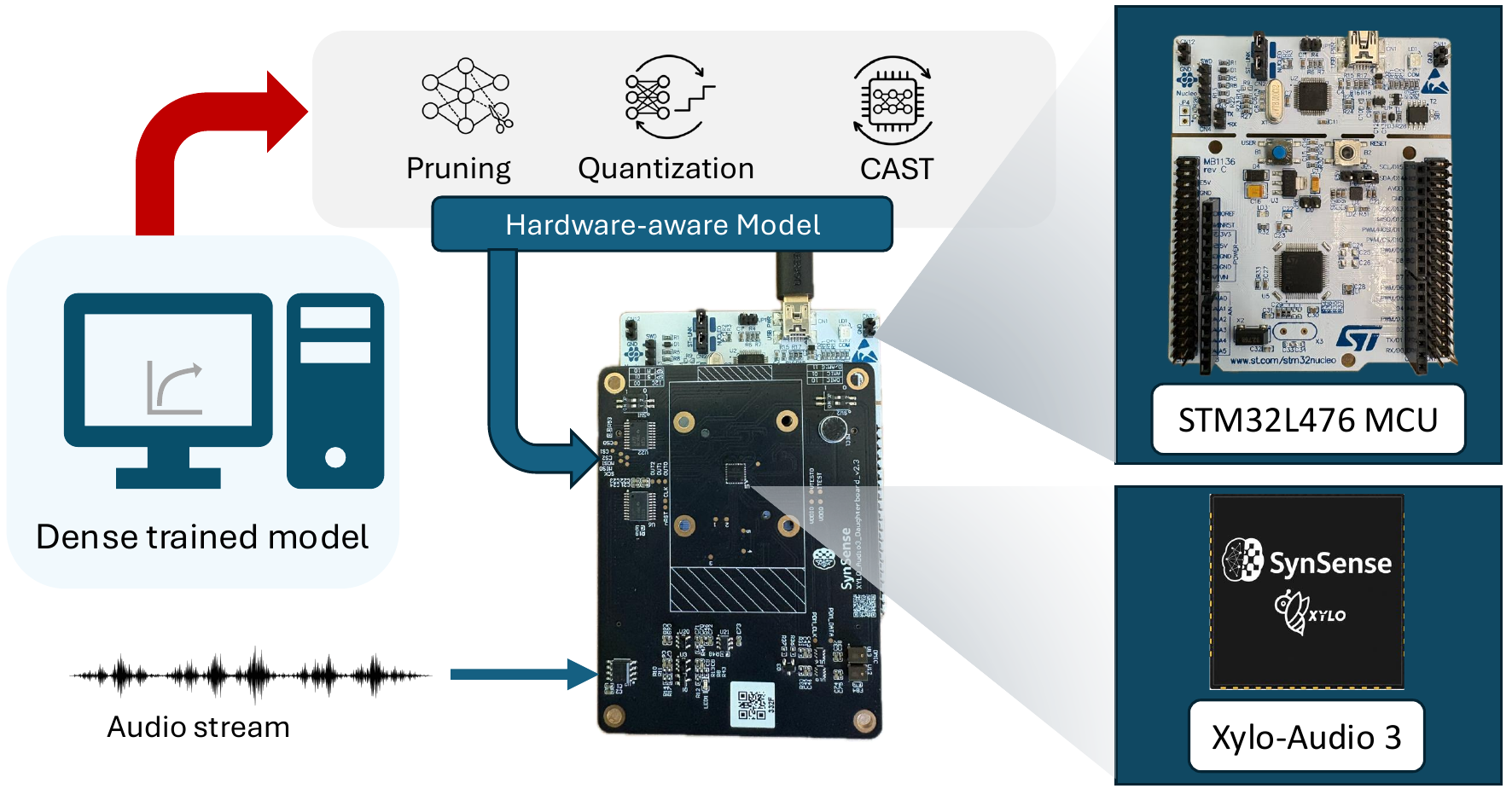}
  \vspace{-10pt}
  \caption{\name hardware: the trained model is mapped to Xylo Audio~3
  (bottom right), which is stacked on an STM32L476
  Nucleo host (top right).}
  \vspace{-12pt}
  \label{fig:hardware}
\end{figure}

Fig.~\ref{fig:hardware} shows the hardware: the Xylo Audio~3 board, running
the mapped CAST backbone and SEE, is stacked on the STM32L476 host.
\textbf{Hybrid manager.} At the end of each three-second window $w$, a cycle
scheduler triggers the readout of the $32$ SEE spike counts
$\mathbf{n}^{(w)} = \sum_{t=1}^{T}\mathbf{r}^{(w)}_t$ over UART, and a mode
controller routes them to identification or enrollment. Each count fits in a
16-bit integer.
\textbf{von Neumann core.} $\mathbf{n}^{(w)}$ is exactly the quantity the SEE
normalizes in training, so the STM32 recovers the trained embedding
$\mathbf{e}$ (Sec.~\ref{sec:see}) by normalizing $\mathbf{n}^{(w)}$,
subtracting the stored centring vector $\boldsymbol{\mu}$, and normalizing
again. To identify, it
averages the embeddings of the latest $W=3$ windows, scores the average
against every prototype by cosine similarity, and accepts the best match only
if its score reaches the threshold for the current gallery size, since the
highest guest score rises as the gallery grows (Sec.~\ref{sec:eval-household}).
To enroll, it accumulates a new speaker's normalized embeddings into one
prototype, stored as $32$ signed 16-bit fixed-point values, $64$ bytes per
speaker. A decision takes fewer than $10^3$
operations at $n=20$; enrollment and deletion change only the gallery and the
threshold in use, and the chip configuration stays fixed.

%% file: Sections/7-evaluation_new.tex
\section{Evaluation}
\label{sec:evaluation}

We evaluate \name as a few-shot open-set recognizer, to simulate the
deployment-time situation in which a user enrolls new speakers from a single
recording by forward passes alone, using audio that was never used to train
\name. Sec.~\ref{sec:eval-method} defines the
datasets, the few-shot protocol, the metrics, and the training budgets.
Sec.~\ref{sec:eval-baselines} compares \name with the baselines on
recognition and system cost.
Sec.~\ref{sec:eval-ablation} isolates the contribution of each HTA design
choice at the deployed neuron and synapse budget. Sec.~\ref{sec:eval-phase1}
reports the HTA-SNN backbone on held-out VoxCeleb2.
Sec.~\ref{sec:eval-see-ablation} isolates the contribution of each SEE design
choice on the frozen backbone. Sec.~\ref{sec:eval-household} evaluates the full \name system, backbone and
SEE, on few-shot recognition in simulated households.
Sec.~\ref{sec:eval-ondevice} describes the three training steps that deploy
\name on Xylo Audio~3 and evaluates the deployed system on the chip.


\subsection{Evaluation Methodology}
\label{sec:eval-method}

\noindent \textbf{Datasets.} Each stage of \name uses a disjoint speaker
population. The backbone is pretrained on VoxCeleb2~\cite{chung2018voxceleb2}
($2.05$M three-second segments, $5994$ speakers) and evaluated on a
per-speaker stratified hold-out of $224{,}760$ segments from the same
speakers. The SEE is pretrained on VoxCeleb1~\cite{nagrani2017voxceleb}
(Sec.~\ref{sec:see-pretraining}), whose $1251$ speakers are absent from
VoxCeleb2. Few-shot recognition is evaluated on Common
Voice~\cite{aardila2020commonvoice}, whose speakers are unseen by both the
backbone and the SEE. All audio passes through a software model of
the Xylo Audio~3 audio front end (AFE), giving a $16$-channel spike
raster. The chip clips this raster to $15$ events per bin, which enters only the
deployment steps of Sec.~\ref{sec:eval-ondevice}.

\noindent \textbf{Few-shot household protocol.} We build $20$ Common Voice
speaker pools of $10$ male and $10$ female speakers who share one accent label,
with no speaker in two pools. Pools are assigned round-robin over
the accents, so they span $8$ accents. Each pool yields one household of every
size $n\in\{2,4,6,8,10,14,18,20\}$ from its first $n/2$ male and $n/2$ female
speakers, so sizes are nested and gender balanced. Each member is enrolled
from their first $60\,\mathrm{s}$ ($29$ windows at a $2\,\mathrm{s}$ stride;
Sec.~\ref{sec:see-gallery}). The next $120\,\mathrm{s}$ yield $40$
non-overlapping query windows. Each decision averages the latest $W=3$
windows in a sliding fashion, so after the first $9\,\mathrm{s}$ a new
decision is issued every $3\,\mathrm{s}$, giving $38$ decisions per speaker. Following openFEAT~\cite{kc2022openfeat}, $50n$ guest
decisions are drawn uniformly at random from speakers outside the household,
regardless of accent, so about $10\%$ of guest decisions come from speakers who
share the household's accent.

\noindent \textbf{Metrics.} All scores are cosine similarities of
L2-normalized embeddings, mean-centred for the SEE
(Sec.~\ref{sec:see-concept}). On households, \emph{top-1} is the fraction of
member decisions assigned to the correct prototype, and \emph{rejection EER}
is the operating point of the rejection threshold on $s_{\max}$, the highest
cosine similarity between a decision and any enrolled prototype
(Sec.~\ref{sec:see-gallery}), at which the fraction of
guests accepted equals the fraction of members rejected; both are averaged over
households. For the backbone, \emph{top-1} assigns each held-out VoxCeleb2
segment to the nearest of the $5994$ normalized AAM-Softmax class vectors
(no margin), measuring separation of the training identities, and
\emph{EER} over $50{,}000$ random pairs of VoxCeleb1 windows ($20$ per
speaker) measures transfer to unseen speakers, the condition under which the
SEE uses the backbone.

\noindent \textbf{Ablation protocol.} We run two ablation studies, one on the
HTA-SNN backbone (temporal stages, skip connections, and synaptic timescales)
and one on the SEE (training objective, centring, and readout weights). Each
study changes one design factor at a time and keeps all others fixed. $\Delta$ is each variant's
difference from its reference, reported with a 95\% confidence interval
(Secs.~\ref{sec:eval-ablation} and~\ref{sec:eval-see-ablation}). The HTA-SNN
ablations vary the backbone, and their reference is the deployed HTA-SNN. Every
backbone variant has $832$ LIF neurons and exactly $30{,}024$ synapses
($29{,}000$ pruned weights plus the unpruned $1024$-weight input projection,
within the $31{,}744$-synapse budget). Every backbone variant is trained with
the full pipeline and epoch budget of the deployed backbone
(Sec.~\ref{sec:eval-phase1}): $1000$ epochs of AAM-Softmax pretraining, the
same SpikeFit pruning schedule~\cite{spikefit2025}, and $30$ epochs of bitshift
QAT. Table~\ref{tab:ablations} gives each variant's
best pretraining and pruning learning rates, and bitshift QAT uses
$3\times10^{-4}$ for every variant. The SEE ablations vary the SEE on the same
frozen HTA-SNN backbone, and their reference is the mean-centred SEE
(Sec.~\ref{sec:see-concept}). Every SEE variant keeps
the $32$-neuron int8 readout and is evaluated on the $151$ VoxCeleb1 test
speakers, and every trained SEE variant follows the pretraining recipe of the
SEE (Secs.~\ref{sec:see-pretraining} and~\ref{sec:eval-see-ablation}).

\noindent \textbf{Baselines.} On households we compare against four
reproduced ANN baselines: the
small-footprint verification network of Balian et
al.~\cite{balian2021smallfootprint} (Small-footprint SV), DS-CNN-S~\cite{zhang2018helloedge},
BC-ResNet-1~\cite{kim2021bcresnet}, and MLP-d128, a dense framewise control
with HTA-SNN's input and hidden width. Each is pretrained on VoxCeleb2 with
AAM-Softmax on its native-width embedding and enrolled and scored like \name.
DS-CNN-S and BC-ResNet-1, keyword-spotting classifiers, take a linear
embedding layer at their pooled width ($64$ and $20$) for this pretraining.
SynNet~\cite{muir2023xylo,bos2024micropower}, the reference SNN for Xylo, is the spiking
baseline for the backbone (Sec.~\ref{sec:eval-phase1}). It tiles a fixed bank
of eight timescales ($2$--$256$) across each layer, has no skip connections,
and is trained with the full pipeline of the deployed backbone at the same
neuron and synapse budget.

\noindent \textbf{System metrics.} ANN baselines run as int8 models on an
STM32U575ZIT6Q~\cite{st2024stm32u575ds} ($2\,\mathrm{MB}$ flash, $786\,\mathrm{KB}$ SRAM) and HTA-SNN
on Xylo Audio~3, both at $50\,\mathrm{MHz}$. The baselines need this larger
memory, since \name's host, the STM32L476, has $128\,\mathrm{KB}$ of SRAM and
$1\,\mathrm{MB}$ of flash~\cite{st2024stm32l476ds}. We report per-window
execution time, idle power with the model loaded, dynamic power above idle,
dynamic energy per decision (dynamic power $\times$ execution time), and total
energy per decision, the sum of the idle and dynamic energy
((idle power $+$ dynamic power) $\times$ execution time).
Execution time covers loading the window's input into the device and the
forward pass. For the baselines, this includes buffering the window's MFCC
features on the MCU, so their execution time and dynamic energy include the
buffering. For \name, execution time is the time Xylo Audio~3 spends on one
window, from transferring the window's AFE spike raster into the chip through
the $293$ timesteps of the network to reading out the $32$ readout spike
counts. It excludes the development kit's USB and host-side overhead, which
is not part of a deployed system. \name's numbers are end to end, adding the
STM32L476 host's $0.16\,\mathrm{ms}$ of cosine scoring per window and its idle
power.
Table~\ref{tab:household-scaling} compares
\name with all baselines on recognition and system cost; Sec.~\ref{sec:eval-baselines}
discusses the results.

\begin{table*}[t]
\centering
\caption{Few-shot household scaling and deployment cost, mean over $20$ accent-homogeneous households.}
\label{tab:household-scaling}
\small
\setlength{\tabcolsep}{4pt}
\renewcommand{\arraystretch}{1.08}
\begin{subtable}{\linewidth}
\centering
\caption{Top-1 identification and rejection EER (\%) with sliding $W{=}3$ decisions; lower EER is better. Size: weights of the baselines, synapses of \name.}
\label{tab:household-recog}
\setlength{\tabcolsep}{3pt}
\begin{tabular}{lllrrrrrrrr}
\toprule
 &  &  & \multicolumn{2}{c}{$n{=}2$} & \multicolumn{2}{c}{$n{=}4$} & \multicolumn{2}{c}{$n{=}6$} & \multicolumn{2}{c}{$n{=}20$} \\
\cmidrule(lr){4-5}\cmidrule(lr){6-7}\cmidrule(lr){8-9}\cmidrule(lr){10-11}
Backbone & Input & Size & Top-1 & EER & Top-1 & EER & Top-1 & EER & Top-1 & EER \\
\midrule
Small-footprint SV~\cite{balian2021smallfootprint}
  & MFCC-40 & $236.6$k & $100.0$ & $0.6$ & $99.7$ & $0.9$ & $99.8$ & $1.3$ & $99.6$ & $2.4$ \\
DS-CNN-S~\cite{zhang2018helloedge}
  & MFCC-40 & $26.5$k & $100.0$ & $1.5$ & $99.5$ & $1.7$ & $99.6$ & $2.4$ & $99.3$ & $4.1$ \\
BC-ResNet-1~\cite{kim2021bcresnet}
  & MFCC-40 & $7.0$k & $99.7$ & $2.8$ & $98.9$ & $3.2$ & $99.0$ & $4.5$ & $97.6$ & $7.7$ \\
MLP-d128
  & MFCC-16 & $146.9$k & $98.8$ & $15.0$ & $93.8$ & $19.4$ & $92.5$ & $22.8$ & $85.6$ & $27.9$ \\
\rowcolor{green!12}\textbf{\name} \textit{(ours)}
  & \textbf{AFE-16} & $\mathbf{29.0}$\textbf{k} & $99.6$ & $5.5$ & $98.8$ & $7.2$ & $98.0$ & $9.3$ & $94.9$ & $13.2$ \\
\bottomrule
\end{tabular}
\end{subtable}

\medskip
\begin{subtable}{\linewidth}
\centering
\caption{Deployment cost per inference: execution time, idle and dynamic power, and dynamic and total energy.}
\label{tab:household-cost}
\begin{tabular}{lrrrrr}
\toprule
Backbone & Exec. time & Idle P & Dyn. P & Dyn. E & Tot. E \\
 & (ms) & (mW) & (mW) & (mJ/inf.) & (mJ/inf.) \\
\midrule
Small-footprint SV~\cite{balian2021smallfootprint}
  & 2149.10 & 0.05 & 13.32 & 28.63 & $28.7$ \\
DS-CNN-S~\cite{zhang2018helloedge}
  & 2022.28 & 0.05 & 13.38 & 27.06 & $27.2$ \\
BC-ResNet-1~\cite{kim2021bcresnet}
  & 1659.71 & 0.05 & 13.26 & 22.01 & $22.1$ \\
MLP-d128
  & 3413.40 & 0.05 & 12.69 & 43.32 & $43.5$ \\
\rowcolor{green!12}\textbf{\name} \textit{(ours)}
  & $\mathbf{961.9}$ & $\mathbf{2.30}$ & $\mathbf{1.70}$ & $\mathbf{1.63}$ & $\mathbf{3.8}$ \\
\bottomrule
\end{tabular}
\end{subtable}
\renewcommand{\arraystretch}{1.0}
\end{table*}


\subsection{Comparison with Baselines}
\label{sec:eval-baselines}

\begin{figure*}[t]
  \centering
  \includegraphics[width=\textwidth]{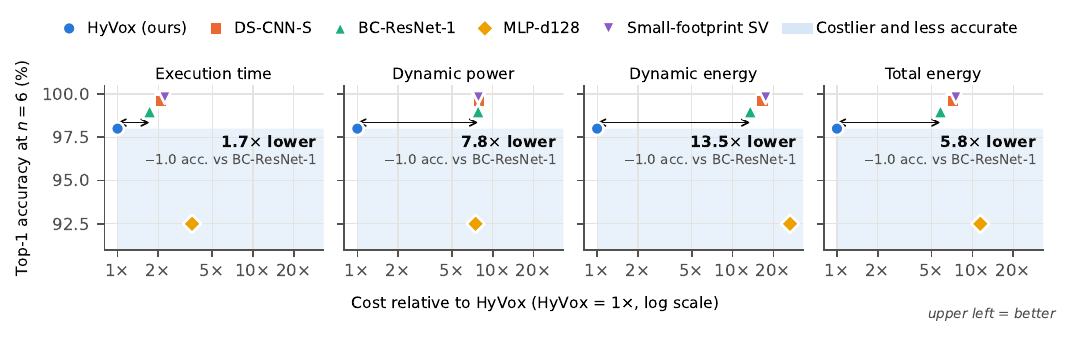}

  \caption{Pareto view of top-1 at $n=6$ against system cost relative to \name, on a log scale. Arrows: cost
  ratio to the cheapest more accurate baseline.}
  \label{fig:pareto-cost}

\end{figure*}

Table~\ref{tab:household-scaling}
compares \name with the four ANN baselines of Sec.~\ref{sec:eval-method} on
both metrics and on system cost.
A shared device typically enrolls few speakers: openFEAT~\cite{kc2022openfeat}
evaluates groups of $2$ to $7$ and treats $10$ as larger than typical. In this
range, at $n=4$ and $n=6$, \name is within $0.9$ and $1.8$ top-1 points of
the strongest baseline, Small-footprint SV, which spends $7.6\times$ the total
energy per decision, and within $0.7$ and $1.6$ points
of DS-CNN-S, which spends $16.6\times$ the dynamic energy and $7.2\times$ the
total energy per decision; its rejection EER is $5.5$ and $6.9$ points above
DS-CNN-S.
Fig.~\ref{fig:pareto-cost} plots top-1 at $n=6$ against each system cost.
Each baseline that is more accurate costs more in execution time and energy,
from $1.7\times$ the execution time to $5.8\times$ the total energy per
decision of \name (both for BC-ResNet-1), for $1.0$ more top-1
points, and MLP-d128 is both costlier and less accurate.
As Table~\ref{tab:household-cost} shows, modern MCUs from ultra-low-power
families reduce idle power to as low as $50\,\mu\mathrm{W}$, a level current
neuromorphic processors cannot reach, so the MCU baselines draw less idle
power than \name ($0.05$ against $2.30\,\mathrm{mW}$). However, running the
SNN on a neuromorphic processor makes up for this in dynamic power, where
\name is $7.5$--$7.9\times$ lower than every baseline. Overall, \name consumes
$5.8$--$11.4\times$ less total energy per decision than the baselines, and
this advantage will grow as the idle power of next-generation neuromorphic
chips is brought down further.

At $n=20$, the compact ANN baselines lead by $2.7$ (BC-ResNet-1) and $4.4$
(DS-CNN-S) top-1 points while spending $13.5\times$ and $16.6\times$ the
dynamic energy and $5.8\times$ and $7.2\times$ the total energy per decision. The
dense MLP-d128 control, with HTA-SNN's input and hidden width, reaches
$85.6\%$ top-1 and $27.9\%$ rejection EER at $n=20$; \name is $9.3$ top-1
points higher and $14.7$ EER points lower, with a backbone of $5.1\times$
fewer weights in integer, bitshift-quantized form and $26.6\times$ less
dynamic energy per decision.


\subsection{HTA-SNN Ablation Studies}
\label{sec:eval-ablation}

The ablations identify which design factors of the HTA-SNN backbone
(Sec.~\ref{sec:hta-architecture}) account for its accuracy: the temporal stages, the
skip connections, and the learned timescales and their initialization. Each
ablation changes one factor, keeps all others fixed, and holds the deployed
neuron and synapse budget (Sec.~\ref{sec:eval-method}).
Table~\ref{tab:ablations} groups the variants by factor below HTA-SNN.
At the shared learning rates, training collapses with linear-uniform
initialization, and pruning collapses without skip~A. For these variants and
the variant without both skips, a learning-rate sweep selects the rates listed
in Table~\ref{tab:ablations}. All three skip variants share the highest
pruning rate that is stable without skip~A, which keeps the skip rows
comparable. $\Delta$ is each variant's difference from HTA-SNN.
To quantify evaluation uncertainty, we resample the test speakers with
replacement $1{,}000$ times and score HTA-SNN and every variant on each
resample. The 95\% confidence interval is $[\Delta_{2.5},\, \Delta_{97.5}]$, the $2.5$th
and $97.5$th percentiles of the $1{,}000$ resampled $\Delta$. Resampling
whole speakers keeps the correlated segments of each speaker together.

\noindent \textbf{Temporal stages.} The stage variants remove one
temporal stage at a time and reassign its neurons to the adjacent stages
(local to medium; medium split between local and global; global to medium),
reattaching both skip connections to the nearest remaining stage. The neuron
count is therefore unchanged, and the widths in parentheses give the resulting
local, medium, and global stage sizes. Every removal degrades both metrics. The global
stage is the most important ($-3.4$ top-1, $+1.3$ EER). Since its neurons
remain in the network, this loss is attributable to the long integration
window itself. Removing the local stage costs $2.0$ top-1 and $0.8$ EER
points, and removing the medium stage costs $0.6$ top-1 and $0.6$ EER
points.

\noindent \textbf{Skip connections.} The skip variants remove skip~A
(input projection to global stage), skip~B (local stage to readout), or both.
Each skip benefits identification: removing skip~A costs $3.4$ top-1 points,
removing skip~B $4.7$, and removing both $6.2$. Skip~A also benefits transfer
to unseen speakers, with EER $0.9$ points higher without it; the EER change of
the other two variants ($+0.1$) lies within its 95\% confidence interval and
thus remains comparable with HTA-SNN.

\noindent \textbf{Learned versus fixed timescales.} The timescale variants
replace the learned per-neuron $\tau_{\mathrm{syn}}$ of HTA-SNN with a single
fixed value shared by all four trainable stages. No fixed value matches the
learned timescales on unseen speakers: every setting raises EER by
$1.8$--$3.6$ points. Fixed timescales also force a trade-off between the two
metrics. The shortest timescale ($\tau_{\mathrm{syn}}=2$) improves top-1 on
the training identities by $0.8$ points, yet raises EER by $2.1$ points, which
indicates overfitting to the pretraining speakers. Longer timescales degrade
both metrics: $\tau_{\mathrm{syn}}=4$ costs $1.4$ top-1 and $1.8$ EER points,
and $\tau_{\mathrm{syn}}=16$ costs $12.2$ top-1 and $3.6$ EER points.
Performance is also highly sensitive to the chosen value, with top-1 varying
by $13.0$ points between $\tau_{\mathrm{syn}}=2$ and $16$. Learned timescales
remove this search and resolve the trade-off: in the deployed backbone, the
median learned $\tau_{\mathrm{syn}}$ ranges from about $2.8$ in the local
stage to about $58$ in the readout (Sec.~\ref{sec:eval-phase1}), a spread
that no single shared value can represent.

\noindent \textbf{$\tau$ initialization.} The $\tau$ initialization variant
draws the initial synaptic time constant $\tau_0$, the value of each neuron's
$\tau_{\mathrm{syn}}$ before training, from a linear-uniform distribution over
$[2,256]$ (median $129$). HTA-SNN draws it from a log-uniform distribution over
the same range (equal mass per octave, median $22.6$). The log-uniform and
linear-uniform distributions place $42.9\%$ and $5.5\%$ of neurons,
respectively, below $\tau_0=16$.
Linear-uniform initialization converges only at an $8\times$ smaller
learning rate and costs $12.3$ top-1 and $2.7$ EER points, the largest top-1
loss in the study.

\begin{table*}[t]
\centering
\caption{Backbone ablations; widths in parentheses: local/medium/global. HTA-SNN row: top-1 and EER in \%; other rows: $\Delta$ from HTA-SNN. $\pm$: 95\% confidence interval over $1{,}000$ resamples of the test speakers.}
\label{tab:ablations}
\small
\renewcommand{\arraystretch}{1.15}
\setlength{\tabcolsep}{5pt}
\begin{tabular}{clcccc}
\toprule
 & & \multicolumn{2}{c}{Best learning rate} & & \\
\cmidrule(lr){3-4}
\makebox[1em]{} & Variant & Pretraining & Pruning & $\Delta$Top-1\,$\uparrow$ & $\Delta$EER\,$\downarrow$ \\
\midrule
\rowcolor{green!12}\multicolumn{2}{l}{\textbf{HTA-SNN}} & $7.9\times10^{-4}$ & $3.0\times10^{-4}$ & $39.6\pm0.7$ & $21.5\pm0.4$ \\
\midrule
\multirow{3}{*}{\rotatebox[origin=c]{90}{\footnotesize Stages}} & w/o local (--/384/256) & $7.9\times10^{-4}$ & $3.0\times10^{-4}$ & $-2.0\pm0.3$ & $+0.8\pm0.3$ \\
 & w/o medium (256/--/384) & $7.9\times10^{-4}$ & $3.0\times10^{-4}$ & $-0.6\pm0.2$ & $+0.6\pm0.3$ \\
 & w/o global (128/512/--) & $7.9\times10^{-4}$ & $3.0\times10^{-4}$ & $-3.4\pm0.2$ & $+1.3\pm0.3$ \\
\midrule
\multirow{3}{*}{\rotatebox[origin=c]{90}{\footnotesize Skips}} & w/o skip A ($\mathbf{h}_1\!\rightarrow\!\mathbf{h}_4$) & $7.9\times10^{-4}$ & $3.0\times10^{-5}$ & $-3.4\pm0.2$ & $+0.9\pm0.3$ \\
 & w/o skip B ($\mathbf{h}_2\!\rightarrow\!\mathbf{h}_5$) & $7.9\times10^{-4}$ & $3.0\times10^{-5}$ & $-4.7\pm0.2$ & $+0.1\pm0.3$ \\
 & w/o skips A and B & $4.0\times10^{-4}$ & $3.0\times10^{-5}$ & $-6.2\pm0.3$ & $+0.1\pm0.3$ \\
\midrule
\multirow{3}{*}{\rotatebox[origin=c]{90}{\footnotesize Timescale}} & Fixed $\tau_{\mathrm{syn}}=2$ & $7.9\times10^{-4}$ & $3.0\times10^{-4}$ & $+0.8\pm0.2$ & $+2.1\pm0.3$ \\
 & Fixed $\tau_{\mathrm{syn}}=4$ & $7.9\times10^{-4}$ & $3.0\times10^{-4}$ & $-1.4\pm0.2$ & $+1.8\pm0.4$ \\
 & Fixed $\tau_{\mathrm{syn}}=16$ & $7.9\times10^{-4}$ & $3.0\times10^{-4}$ & $-12.2\pm0.3$ & $+3.6\pm0.4$ \\
\midrule
\rotatebox[origin=c]{90}{\footnotesize $\tau$ init.} & Linear-uniform $\tau_0$\rule[-2.2ex]{0pt}{5.6ex} & $9.9\times10^{-5}$ & $7.5\times10^{-5}$ & $-12.3\pm0.3$ & $+2.7\pm0.4$ \\
\bottomrule
\end{tabular}
\end{table*}


\subsection{HTA-SNN Pretraining and Evaluation}
\label{sec:eval-phase1}

The deployed backbone is the HTA-SNN of Sec.~\ref{sec:hta-architecture},
trained with AAM-Softmax~\cite{deng2019arcface} ($m=0.2$, $s=24$) on all
$5994$ VoxCeleb2 identities.

\begin{figure}[h]
  \centering

  \begin{minipage}[t]{0.49\linewidth}
    \centering
    \includegraphics[width=\linewidth]{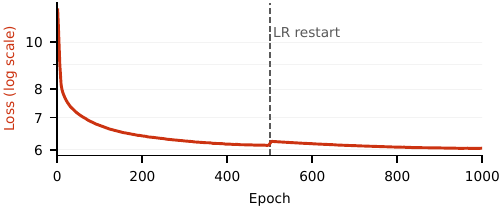}\\
    {\small (a) AAM-Softmax loss.}
  \end{minipage}\hfill
  \begin{minipage}[t]{0.49\linewidth}
    \centering
    \includegraphics[width=\linewidth]{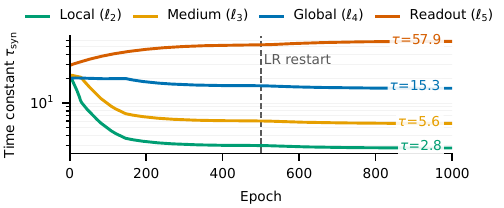}\\
    {\small (b) Per-stage $\tau_{\mathrm{syn}}$.}
  \end{minipage}

  \caption{HTA-SNN backbone training.}

  \label{fig:spikefit-training}
\end{figure}

\noindent \textbf{Training dynamics.} Fig.~\ref{fig:spikefit-training} traces
$1000$ dense epochs in two cosine-annealed runs with a warm restart at
epoch~$500$. Panel~(b) shows the per-stage median
$\tau_{\mathrm{syn}}=\exp(\overline{\log\tau})$ separating into the intended
hierarchy: the readout is slowest ($\tau\approx58$), the local stage fastest
($\tau\approx2.8$), and the medium and global stages lie between.

\noindent \textbf{Quantitative results.} Table~\ref{tab:phase1-results}
compares the HTA-SNN backbone with SynNet~\cite{muir2023xylo,bos2024micropower},
both compressed to the same $29.0$k synapses and quantized for the chip's
integer dynamics (Sec.~\ref{sec:eval-ondevice}). HTA-SNN leads by $4.1$, $4.2$, and
$3.9$ points of top-1, top-5, and top-10 and lowers the EER by $1.6$ points.

\begin{table}[t]
\centering
\caption{Backbones (\%): top-$k$ on held-out VoxCeleb2, EER on unseen VoxCeleb1 speakers (Sec.~\ref{sec:eval-method}). Syn.: synapses, excluding the $1$k-weight input projection.}
\label{tab:phase1-results}
\small
\setlength{\tabcolsep}{4pt}
\begin{tabular*}{0.9\linewidth}{@{\extracolsep{\fill}}lrrrrr}
\toprule
Model & Syn. & Top-1 & Top-5 & Top-10 & EER \\
\midrule
\textbf{HTA-SNN} \textit{(ours)} & $29.0$k  & $\mathbf{39.6}$ & $\mathbf{60.5}$ & $\mathbf{68.8}$ & $\mathbf{21.5}$ \\
SynNet~\cite{muir2023xylo,bos2024micropower} & $29.0$k & $35.5$ & $56.3$ & $64.9$ & $23.1$ \\
\bottomrule
\end{tabular*}
\end{table}


\subsection{SEE Ablation Studies}
\label{sec:eval-see-ablation}

The SEE ablations identify which design factors of the SEE
(Sec.~\ref{sec:see}) account for its accuracy: the training objective, the
centring, and the way the readout weights are obtained. Every variant is
trained and evaluated on top of the same frozen HTA-SNN backbone
(Sec.~\ref{sec:backbone}), so only the SEE and its training change. Each
ablation changes one factor and keeps all others fixed, including the
$32$-neuron int8 readout and the centring of every embedding unless centring
is the factor. Every trained variant follows the pretraining recipe of
the SEE (Sec.~\ref{sec:see-pretraining}), and every checkpoint is selected on
the VoxCeleb1 validation speakers. Table~\ref{tab:see-ablations} reports top-1
identification and open-set EER on the $151$ VoxCeleb1 test speakers, which no
variant sees during training or selection, over $500$ episodes in the
pretraining format (Sec.~\ref{sec:see-pretraining}). Each episode enrolls
$20$ speakers from $4$ windows each, identifies $5$ further windows per
enrolled speaker, and rejects $5$ windows from each of $5$ unenrolled
speakers. $\Delta$ is each variant's difference from the SEE.
To quantify evaluation uncertainty, we resample the $500$ test episodes with
replacement $1{,}000$ times and score the SEE and every variant on each
resample. The 95\% confidence interval is $[\Delta_{2.5},\, \Delta_{97.5}]$, the $2.5$th
and $97.5$th percentiles of the $1{,}000$ resampled $\Delta$. Resampling
whole episodes keeps the enrolled and unenrolled speakers of each episode together.

\noindent \textbf{Training objective.} The objective variants remove the
rejection loss ($\lambda=0$) or replace the episodic objective with closed-set
AAM-Softmax over the $900$ training speakers, whose class weights are
discarded after training. Removing $\mathcal{L}_{\mathrm{rej}}$ costs $2.2$
top-1 and $1.8$ EER points. Closed-set training costs $5.7$ top-1 and $6.0$
EER points, the largest EER loss among the trained variants. A readout trained
to separate a fixed set of speakers therefore transfers poorly to speakers
enrolled from a few windows.

\noindent \textbf{Centring.} The centring variant scores the normalised spike
count $\bar{\mathbf{r}}$ directly, with $\boldsymbol{\mu}=\mathbf{0}$ in
training and inference. Without centring, top-1 drops by $3.9$ points and EER
rises by $2.3$ points. Removing the component that all non-negative
spike-count embeddings share improves both identification and rejection.

\noindent \textbf{Readout weights.} The readout variants keep the readout and
the centring and change how the readout's $32\times128$ projection matrix
$\mathbf{W}_{s}$ (Sec.~\ref{sec:see-concept}), which maps the backbone's
$128$ output spikes to the $32$ readout neurons, is obtained. An untrained readout
at its random initialization costs $20.7$ top-1 and $7.5$ EER points.
The principal component analysis (PCA)~\cite{pearson1901lines} and linear
discriminant analysis (LDA)~\cite{fisher1936use, dehak2011frontfactor}
variants set $\mathbf{W}_{s}$ in closed form, with no episodic training. Each
computes $32$ projection directions from the backbone's mean spike rates over
the windows of the $900$ training speakers and loads them as the rows of $\mathbf{W}_{s}$ in the same int8 readout, scaled by
one input gain chosen on the VoxCeleb1 validation speakers. PCA keeps the
directions of largest variance without using speaker labels, and LDA keeps the
directions that maximise the ratio of between-speaker to within-speaker
variance. Both recover part of the loss of the untrained readout. PCA costs
$8.6$ top-1 and $3.3$ EER points, and LDA costs $7.1$ top-1 and $1.3$ EER
points. Initializing $\mathbf{W}_{s}$ from PCA
before episodic training costs $5.6$ top-1 and $5.1$ EER points, so the
episodic objective reaches a better readout from a random start.

\begin{table}[t]
\centering
\caption{SEE ablations on the VoxCeleb1 test speakers. SEE row: top-1 and EER in \%; other rows: $\Delta$ from the SEE. $\pm$: 95\% confidence interval over $1{,}000$ resamples of the test episodes.}
\label{tab:see-ablations}
\small
\renewcommand{\arraystretch}{1.15}
\setlength{\tabcolsep}{5pt}
\begin{tabular}{clcc}
\toprule
\makebox[1em]{} & Variant & $\Delta$Top-1\,$\uparrow$ & $\Delta$EER\,$\downarrow$ \\
\midrule
\rowcolor{green!12}\multicolumn{2}{l}{\textbf{SEE}} & $59.5\pm0.5$ & $35.3\pm0.6$ \\
\midrule
\multirow{2}{*}{\rotatebox[origin=c]{90}{\footnotesize Obj.}} & w/o $\mathcal{L}_{\mathrm{rej}}$ ($\lambda=0$) & $-2.2\pm0.4$ & $+1.8\pm0.4$ \\
 & Closed-set AAM-Softmax & $-5.7\pm0.4$ & $+6.0\pm0.6$ \\
\midrule
\rotatebox[origin=c]{90}{\footnotesize Cent.} & w/o centring ($\boldsymbol{\mu}=\mathbf{0}$)\rule[-2.2ex]{0pt}{5.6ex} & $-3.9\pm0.4$ & $+2.3\pm0.5$ \\
\midrule
\multirow{4}{*}{\rotatebox[origin=c]{90}{\footnotesize Readout}} & Untrained $\mathbf{W}_{s}$ & $-20.7\pm0.5$ & $+7.5\pm0.6$ \\
 & PCA projection & $-8.6\pm0.4$ & $+3.3\pm0.5$ \\
 & LDA projection & $-7.1\pm0.4$ & $+1.3\pm0.5$ \\
 & PCA init.\ + training & $-5.6\pm0.4$ & $+5.1\pm0.6$ \\
\bottomrule
\end{tabular}
\end{table}


\subsection{Few-Shot Speaker Recognition with \name}
\label{sec:eval-household}

\name combines the deployed HTA-SNN backbone
(Sec.~\ref{sec:eval-phase1}) with the SEE pretrained on VoxCeleb1
(Sec.~\ref{sec:see-pretraining}). The members of each household are enrolled
by forward passes of their enrollment speech through \name, which yield one
prototype per member (Sec.~\ref{sec:see-enrollment}), under the protocol of
Sec.~\ref{sec:eval-method}. We report identification as the household grows,
rejection of unseen speakers, and the embedding geometry behind both.

\begin{figure}[t]
  \centering
  \includegraphics[width=\linewidth]{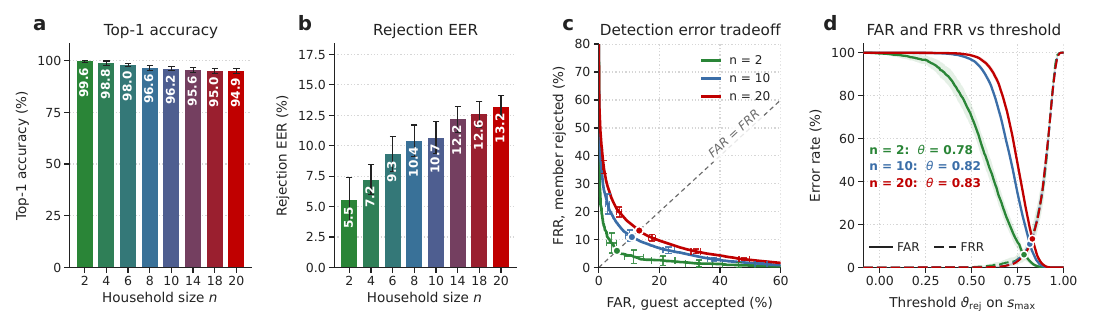}

  \caption{Household identification and unseen-speaker rejection by \name.
  (a)~Top-1 identification; (b)~rejection EER; (c)~detection error tradeoff;
  (d)~FAR (solid) and FRR (dashed) against $\vartheta_{\mathrm{rej}}$. Error
  bars, crosses, and bands are 95\% confidence intervals over $20$
  households; dots mark the EER.}

  \label{fig:household}
\end{figure}

\noindent \textbf{Household scaling.} Fig.~\ref{fig:household}\textbf{a} shows
top-1 identification over all eight household sizes; we report the mean over
the $20$ households with its 95\% confidence interval. Top-1 falls from
$99.6\pm0.6\%$ at $n=2$ to $94.9\pm1.2\%$ at $n=20$. The decline
flattens as the household grows: the six members added from $n=2$ to $n=8$
lower top-1 by $3.0$ points, to $96.6\pm1.2\%$, and the twelve added from
$n=8$ to $n=20$ lower it by a further $1.7$ points. The interval stays within
$\pm1.3$ points at every size.

\noindent \textbf{Unseen-speaker rejection.}
Fig.~\ref{fig:household}\textbf{b}--\textbf{d} evaluate the open-set decision against guests.
Rejection EER rises from $5.5\pm1.8\%$ at $n=2$ to $13.2\pm0.9\%$ at
$n=20$ (Fig.~\ref{fig:household}\textbf{b}, mean and 95\% confidence
interval over households). A member is
identified correctly as long as the correct prototype outscores the other
$n-1$. A guest is rejected only if its similarity to every one of the $n$
prototypes stays below the threshold. The guest score $s_{\max}$ is a
maximum over $n$ cosines and grows with $n$. Because the operating
threshold depends on the application~\cite{tan2021sharedDevices},
Fig.~\ref{fig:household}\textbf{c} plots the detection error tradeoff over all thresholds,
which gives the false reject rate (FRR) at any false accept rate (FAR). At
every FAR, FRR increases with $n$, so a larger household raises the error at
every operating point.
Fig.~\ref{fig:household}\textbf{d} plots both error rates
against the threshold $\vartheta_{\mathrm{rej}}$. The EER threshold moves from
$0.78$ at $n=2$ to $0.83$ at $n=20$, and the FAR curve steepens and shifts
right as $n$ grows. A threshold fixed at the $n=2$ EER point therefore admits
$32.1\%$ of guests at $n=20$, while rejecting $5.8\%$ of members, so the host
sets $\vartheta_{\mathrm{rej}}$ according to the number of enrolled speakers.

\begin{figure*}[t]
  \centering
  \includegraphics[width=\textwidth]{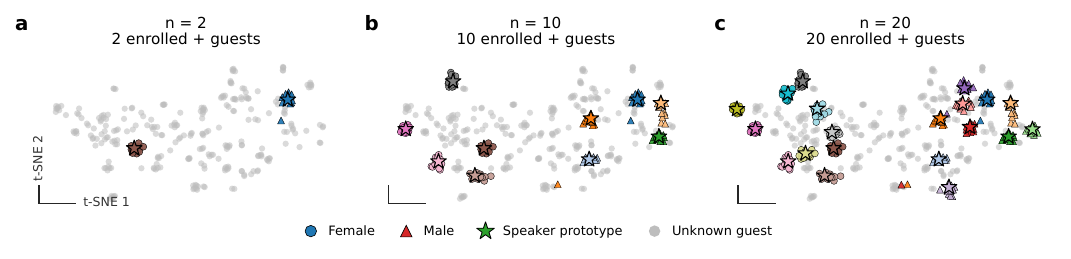}

  \caption{t-SNE of \name embeddings for one household at $n=2$, $10$, and
  $20$ enrolled speakers. Colour identifies the speaker, stars are prototypes,
  and grey points are guest decisions.}

  \label{fig:tsne-household}
\end{figure*}

\noindent \textbf{Embedding geometry.} Fig.~\ref{fig:tsne-household}
projects the \name embeddings of one household (United States English) with a
single t-SNE~\cite{vandermaaten2008tsne} fitted over all points; the three panels differ only in how many
speakers are enrolled, and the guest sample is identical across them. The
decisions of each member form a compact cluster around its prototype at every
household size, consistent with the top-1 of
Fig.~\ref{fig:household}\textbf{a}. At $n=2$ the two prototypes lie at the
edges of the guest cloud. At $n=20$ the prototypes cover much of the region
the guests occupy, so a guest decision lies nearer its closest prototype. This
is the growth of the guest $s_{\max}$ that raises the EER threshold in
Fig.~\ref{fig:household}\textbf{d}. The projection is qualitative, and the
quantitative distances are those reported in Fig.~\ref{fig:household}\textbf{b}--\textbf{d}.

\subsection{On-Device Evaluation}
\label{sec:eval-ondevice}

\begin{figure*}[t]
  \centering
  \includegraphics[width=\textwidth]{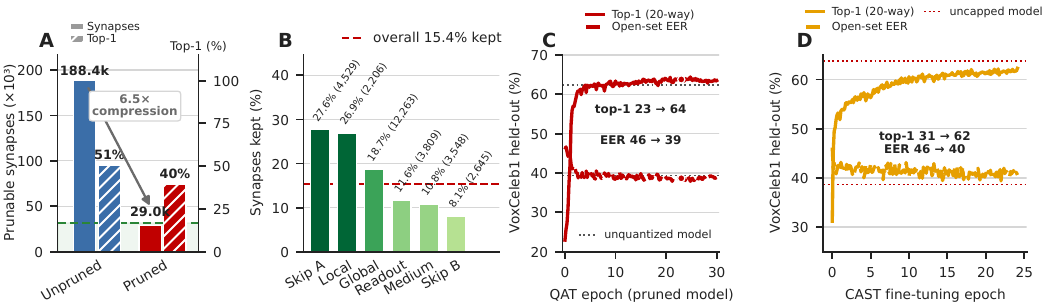}

  \caption{Deploying the HTA-SNN backbone on Xylo Audio~3. (A)~Synapse
  compression; hatched bars: VoxCeleb2 top-1. (B)~Synapses kept per component.
  (C)~Bitshift QAT and (D)~CAST fine-tuning on held-out VoxCeleb1
  speakers; dotted lines: the model from the previous step.}

  \label{fig:deploy-steps}
\end{figure*}

\begin{figure*}[t]
  \centering
  \includegraphics[width=\textwidth]{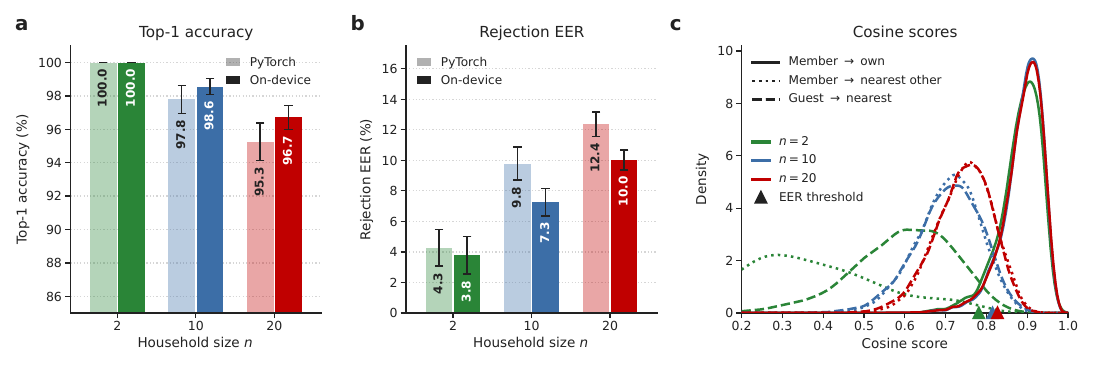}

  \caption{\name on Xylo Audio~3. (a)~Top-1 identification, (b)~unseen-speaker
  rejection EER, and (c)~on-device cosine scores.}

  \label{fig:ondevice}
\end{figure*}

\name runs on Xylo Audio~3 chip with an STM32L476 host. Fitting the network to
the chip takes three training steps after pretraining: synapse compression to the chip's weight budget,
quantization-aware training (QAT) of the weights and timescales for the chip's
integer dynamics, and CAST fine-tuning under the chip's event limits.
Fig.~\ref{fig:deploy-steps} reports the three steps, and
Fig.~\ref{fig:ondevice} the household evaluation of the deployed system.
Every step trains on VoxCeleb2; we measure each step on held-out VoxCeleb1
speakers, who are unseen in training, and report VoxCeleb2 only for the
synapse-compression schedule.

\noindent \textbf{Synapse compression.} The dense backbone has $188.4$k
prunable synapses, $5.9\times$ the $31.7$k budget of the chip's recurrent
weight matrix. We prune it with SpikeFit~\cite{spikefit2025} in $14$ steps
over $1{,}255$ epochs of compression-aware fine-tuning, which leaves $29.0$k
synapses ($15.4\%$, $6.5\times$ compression)
(Fig.~\ref{fig:deploy-steps}\textbf{A}). After the QAT below, the compressed
backbone reaches $39.6\%$ top-1 on held-out VoxCeleb2, against $50.7\%$ for
the dense backbone. Pruning is global i.e. each step keeps the top-scoring
synapses across all stages and skips together, with no per-layer quota, so the
kept share differs by component (Fig.~\ref{fig:deploy-steps}\textbf{B}): skip~A ($27.6\%$) and the local stage
($26.9\%$) keep nearly twice the overall rate, the global stage keeps
$18.7\%$, and the readout, medium stage, and skip~B keep $11.6\%$ or less.

\noindent \textbf{QAT of weights and timescales.} The chip stores weights as
integers and realizes every synaptic decay as a bit shift. Compression
fine-tuning trains the weights against their quantization codebooks
(Sec.~\ref{sec:compression}), and before
the last pruning step every $\tau_{\mathrm{syn}}$ is snapped to the
power-of-two grid and frozen. Bitshift QAT
then replaces the floating-point LIF update with the chip's integer bitshift
update and trains the network for $30$ epochs on VoxCeleb2.
Fig.~\ref{fig:deploy-steps}\textbf{C} tracks the 20-way top-1
(Sec.~\ref{sec:see-pretraining}) and open-set EER of the backbone embedding on held-out VoxCeleb1 validation speakers. On
the $151$ VoxCeleb1 test speakers, used neither for training nor for
checkpoint selection (Table~\ref{tab:cast}), integer dynamics without QAT cost
$37.2$ top-1 and $9.3$ EER points against the unquantized model, and the QAT model
ends $1.2$ top-1 points above it and $1.0$ EER point below.

\noindent \textbf{CAST fine-tuning.} Fig.~\ref{fig:deploy-steps}\textbf{D}
tracks CAST fine-tuning (Sec.~\ref{sec:cast}) on the VoxCeleb1 validation
speakers, where it recovers most of the loss of mapping the QAT backbone
directly to the chip. On the test speakers (Table~\ref{tab:cast}), the directly mapped
backbone loses $29.9$ top-1 and $6.4$ EER points against the unquantized model,
and after CAST it differs from it by $0.4$ and $0.1$ points, both within their
95\% confidence intervals.

The SEE is then fine-tuned on the chip's backbone spikes
(Sec.~\ref{sec:chip-readout}), with the centring vector $\boldsymbol{\mu}$
re-estimated from the chip's readout counts. On the CAST backbone, the
pretrained SEE without this fine-tuning reaches $42.3\%$ top-1 and $43.1\%$
EER on the VoxCeleb1 validation speakers; fine-tuning raises top-1 to
$58.5\%$ and lowers the EER to $36.0\%$.
These VoxCeleb1 errors are far higher than the household errors of
Sec.~\ref{sec:eval-household} because the VoxCeleb1 episodes are a much harder
protocol. Each episode enrolls $20$ speakers from only $4$ windows each and
scores every $3\,\mathrm{s}$ query window on its own, and the windows come from
in-the-wild interviews whose channel and background noise vary within a
speaker. A household member is enrolled from $29$ windows of read Common Voice
speech, and each decision averages $W=3$ windows. The hardness of the
VoxCeleb1 episodes, used both to train and to evaluate the SEE, is
deliberate. It keeps accuracy far from saturation, so the cost of each chip
constraint and the gain of each training step remain measurable. The Common
Voice households relax these conditions to those of a practical deployment and
measure the accuracy of the deployed system.

\begin{table}[t]
\centering
\caption{Deployment steps of the HTA-SNN backbone on VoxCeleb1 test speakers
(\%). HTA-SNN row: pruned, unquantized model in PyTorch; other rows: $\Delta$
from it. $\pm$: 95\% confidence interval over $1{,}000$ resamples of the test episodes.}
\label{tab:cast}
\small
\renewcommand{\arraystretch}{1.15}
\setlength{\tabcolsep}{5pt}
\begin{tabular}{lrr}
\toprule
Model & $\Delta$Top-1\,$\uparrow$ & $\Delta$EER\,$\downarrow$ \\
\midrule
\textbf{HTA-SNN} (PyTorch, unquantized) & $61.0\pm0.7$ & $38.6\pm1.1$ \\
\midrule
\multicolumn{3}{@{}l}{\textit{Integer bitshift dynamics, in PyTorch}} \\
\quad Without QAT & $-37.2\pm0.8$ & $+9.3\pm1.3$ \\
\rowcolor{green!12}\quad \textbf{With bitshift QAT} & $\mathbf{+1.2\pm0.4}$ & $\mathbf{-1.0\pm0.4}$ \\
\midrule
\multicolumn{3}{@{}l}{\textit{Bitshift-QAT model on the chip}} \\
\quad Mapped directly & $-29.9\pm0.7$ & $+6.4\pm1.1$ \\
\rowcolor{green!12}\quad \textbf{After CAST} ($26$ epochs) & $\mathbf{-0.4\pm0.5}$ & $\mathbf{+0.1\pm0.6}$ \\
\bottomrule
\end{tabular}

\end{table}

\noindent \textbf{Why the chip needs CAST.} Mapped directly, the bitshift-QAT
model loses $31.1$ top-1 points and gains $7.4$ EER points against its own
PyTorch accuracy on the test speakers. To locate this loss, we replicate the
chip's neuron update in PyTorch with each chip constraint switchable. With
every constraint off, the replica matches the PyTorch model, and with every
constraint on, it matches Xylo Audio~3. We then add
the constraints to the trained model one at a time (Table~\ref{tab:chip-loss}).
Four constraints each cost $18$ to $32$ top-1 points on their own. The chip
caps every neuron at $31$ output events per time step. It stores weights as
8-bit integers (int8) with one scale per destination layer, so a few large
weights set the scale and round many small weights to zero, and the skip
paths, folded into the same recurrent matrix, share that scale. Every
hidden-to-hidden connection adds a one-step delay. Finally, the trained
floating-point thresholds are mapped to the chip's integer thresholds. Bitshift QAT trains the network for the chip's
bit-shift decays; none of these constraints is part of its training model.

\begin{table}[t]
\centering
\caption{CAST ablation on VoxCeleb1 test speakers (\%). HTA-SNN row: PyTorch
after bitshift QAT; other rows: $\Delta$ from it, each + row adding one chip
constraint. $\pm$: 95\% confidence interval over $1{,}000$ resamples of the test episodes.}
\label{tab:chip-loss}
\small
\renewcommand{\arraystretch}{1.15}
\setlength{\tabcolsep}{5pt}
\begin{tabular}{lrr}
\toprule
Model & $\Delta$Top-1\,$\uparrow$ & $\Delta$EER\,$\downarrow$ \\
\midrule
\textbf{HTA-SNN} (PyTorch, after bitshift QAT) & $62.2\pm0.7$ & $37.5\pm1.1$ \\
\midrule
+ $31$-event cap & $-31.5\pm0.7$ & $+7.7\pm0.9$ \\
+ chip int8 grid & $-26.6\pm0.7$ & $+8.2\pm0.9$ \\
+ hop delay & $-19.3\pm0.7$ & $+7.2\pm0.9$ \\
+ chip thresholds & $-18.2\pm0.6$ & $+6.5\pm0.9$ \\
\midrule
On chip (all constraints) & $-31.1\pm0.8$ & $+7.4\pm1.0$ \\
\rowcolor{green!12}On chip, after \textbf{CAST} & $\mathbf{-1.6\pm0.5}$ & $\mathbf{+1.1\pm0.5}$ \\
\bottomrule
\end{tabular}
\end{table}

Of the four constraints, only the event caps can be imposed during standard
training, through the $31$-event cap, the $15$-event input clamp, and a
cap-aware surrogate gradient. We retrained the whole pipeline from random
initialization with them. In PyTorch this model reaches $59.1\%$ top-1, and on
the chip it falls to $30.0\%$, no better than the directly mapped QAT model,
because the int8 grid, the threshold mapping, and the hop delays remain. With
per-neuron int8 scales and its trained thresholds, it still reaches only
$46.8\%$ top-1 on the chip. Training with the caps thus recovers little of the
chip's loss, so the network has to be trained through all four constraints
together.

CAST trains through all of these constraints together. After $26$ epochs it
comes within $1.6$ top-1 and $1.1$ EER points of the PyTorch model on the
chip, confirming on the test speakers the validation result of
Fig.~\ref{fig:deploy-steps}\textbf{D}, and it exceeds the model trained with
caps by $30.7$ top-1 and $8.3$ EER points with the standard mapping and by
$13.9$ and $3.8$ points with the per-neuron mapping.
CAST trains one integer firing threshold per neuron, which acts as a
per-neuron gain. Across neurons, the thresholds span $3$ to $16$ in the
directly mapped model and $1$ to $142$ after CAST. Strongly driven neurons
receive high thresholds and fire below the $31$-event cap, and weakly driven
neurons receive low ones, so the share of active steps at the cap falls from
$98.5\%$ to $3.7\%$ and graded counts reach the readout. CAST is thus the step that
makes the backbone deployable, recovering $29.5$ of the $31.1$ top-1 points
that the chip's constraints cost.

\noindent \textbf{Hardware deployment.} The deployed network consists of the
CAST HTA-SNN and the SEE, stored as int8 weights, integer thresholds,
and power-of-two decay shifts. We place these parameters into the chip's
input, recurrent, and readout weight matrices, which gives the configuration
that is written to Xylo Audio~3. The synapse counts in this paper, including
the $29.0$k of the compressed backbone, refer to the trained network, where
each skip connection is a separate projection layer. The chip has no skip
layers, so each skip projection is multiplied into the layer it feeds and
stored as direct recurrent synapses, which adds synapses on the skip paths,
while int8 rounding sets the smallest weights to zero. The chip configuration
therefore holds $27{,}259$ recurrent synapses, within its budget of
$31{,}744$. One configuration serves every household, since \name holds no
household-specific weight. Each $3\,\mathrm{s}$ window enters the chip as its
AFE spike raster, starts from a reset state, and returns the $32$ readout
spike counts, and the host forms prototypes and decisions as in
Sec.~\ref{sec:host-inference}. We evaluate $20$ households of $10$ female and $10$ male speakers in four accent
groups, which share no speaker with the households of
Sec.~\ref{sec:eval-household}, with the enrollment, query, guest, and
sliding $W=3$ decision protocol of Sec.~\ref{sec:eval-method}. As the reference, we evaluate the PyTorch model of
Sec.~\ref{sec:eval-household} (Fig.~\ref{fig:household}), the bitshift-QAT
backbone with the pretrained SEE and without the chip's event caps, on these
households.

Fig.~\ref{fig:ondevice} shows that the deployed system on the chip matches or
exceeds the PyTorch model. In Fig.~\ref{fig:ondevice}\textbf{a} and
\textbf{b}, light bars give the PyTorch model and solid bars the on-device
system, with 95\% confidence intervals over the $20$ households (reported as
$\pm$). Top-1 is $100.0\pm0.0\%$, $98.6\pm0.5\%$, and $96.7\pm0.7\%$ at
$n=2$, $10$, and $20$ (Fig.~\ref{fig:ondevice}\textbf{a}), against
$100.0\pm0.0\%$, $97.8\pm0.8\%$, and $95.3\pm1.1\%$ for the PyTorch model.
The rejection EER is $3.8\pm1.2\%$, $7.3\pm0.9\%$, and $10.0\pm0.7\%$
(Fig.~\ref{fig:ondevice}\textbf{b}), against $4.3\pm1.2\%$, $9.8\pm1.1\%$,
and $12.4\pm0.8\%$. Paired over the $20$ households, the two systems agree
within their 95\% confidence intervals at $n=2$ and in top-1 at $n=10$. At
$n=20$ the chip is $1.4\pm0.8$ points higher in top-1, and its EER is
$2.5\pm1.1$ and $2.4\pm0.7$ points lower at $n=10$ and $20$. This gain comes from the backbone trained through CAST. Although the CAST backbone trails the
PyTorch backbone on the VoxCeleb1 test speakers (Table~\ref{tab:cast}), it leads
on the Common Voice households, whose read speech and home recording conditions
differ from the interview recordings of the VoxCeleb2 training data. Training
through the chip's quantization and event limits acts as a regularizer, so the
backbone generalizes better to recording conditions it never saw in training.
Fig.~\ref{fig:ondevice}\textbf{c} pools the on-device cosine scores over
households for members against their own prototype (solid), members against
the nearest other prototype (dotted), and guests against the nearest prototype
(dashed), and a triangle marks the pooled EER threshold of each household size.
The EER threshold moves from $0.78$ at $n=2$ to $0.83$ at $n=20$, the same
upward shift with household size
as in Fig.~\ref{fig:household}\textbf{d}, so the gallery-size-aware threshold
table of Sec.~\ref{sec:host-inference} carries over to the chip.

%% file: Sections/8-related_work.tex
\section{Related Work}
\label{sec:related-work} 

Low-power audio inference rests on mature microcontroller and neuromorphic
baselines, which include compact and end-to-end
benchmarks~\cite{warden2018speechcommands,yik2025neurobench}, efficient
keyword-spotting backbones~\cite{zhang2018helloedge,kim2021bcresnet}, deployment
toolchains~\cite{david2021tflmicro,banbury2021mlperf,stcubeai}, and on-device
speaker identification~\cite{balian2021smallfootprint,gallacher2023speakerid}.
Neuromorphic processors run such workloads with sparse, event-driven
computation~\cite{akopyan2015truenorth,davies2018loihi,furber2014spinnaker,moradi2018dynapse,davies2021loihi}.
Two such chips have been applied to audio. Intel's Loihi is a research chip with
many neuromorphic cores and on-chip
learning~\cite{davies2018loihi,davies2021loihi}, and keyword spotting and signal
processing on it use far less energy per inference than on conventional
processors~\cite{blouw2019benchmarking,blouw2020event}. SynSense's Xylo is a
family of low-power inference chips for edge sensing, and Xylo Audio~3 pairs an
analog audio front-end with a digital core of $1{,}000$ spiking
neurons~\cite{muir2023xylo,synsense2023xyloaudio3ds}, running audio applications
at sub-milliwatt power~\cite{muir2023xylo,bos2024micropower}. These systems have
traditionally been built to recognize a set of labels fixed at training time, so
adding a class means collecting its data and retraining the network. Modern-day
speaker recognition systems have moved from GMMs and i-vectors to neural
embeddings~\cite{reynolds2000gmmubm,dehak2011frontfactor,variani2014dvector,snyder2018xvectors},
and strong
encoders~\cite{okabe2018attentive,desplanques2020ecapa,chen2022wavlm,wang2023campplus,chen2023eres2net}
are trained on large datasets with over $7{,}000$ speakers and over a million
utterances~\cite{nagrani2017voxceleb,chung2018voxceleb2}, using metric
objectives such as GE2E, angular margins, and supervised contrastive
learning~\cite{wan2018ge2e,deng2019arcface,khosla2020supcon}. An encoder maps
each utterance to a fixed-length embedding, and speakers are compared by the
cosine similarity of their embeddings, so a new speaker can be enrolled from a
recording without retraining. Few-shot learning builds a class representation
from a handful of examples, often as the mean embedding of the
class~\cite{snell2017prototypical}, and open-set recognition rejects inputs from
classes that were absent at training
time~\cite{scheirer2013open,geng2021opensetsurvey,vaze2022openset}. Few-shot
open-set learning combines the two and trains on episodes that each enroll new
classes and reject unknown ones~\cite{liu2020peeler}, and openFEAT applies it to
speaker identification in households that share a
device~\cite{kc2022openfeat,tan2021sharedDevices}. Spiking neurons integrate
their input over time in a membrane state and communicate through binary events,
which suits sparse, low-power hardware. LIF dynamics~\cite{gerstner2002spiking}
and surrogate
gradients~\cite{neftci2019surrogate,zenke2018superspike,bellec2018long,shrestha2018slayer,wu2018stbp}
make deep SNNs trainable, and residual, hybrid, and transformer
designs~\cite{lee2020spikebackprop,rathi2020hybrid,fang2021sewresnet,deng2022tet,zhou2023spikformer,zhou2024qkformer}
scale them to larger tasks, beyond the neuron and synapse budget of a
neuromorphic chip. Large sparse networks outperform small dense ones of the same
size~\cite{zhu2017prune}, SpikeFit prunes SNN channels for neuromorphic
deployment and was evaluated on an image classifier that runs for four
timesteps~\cite{spikefit2025}, and quantization-aware training adapts networks
to integer arithmetic~\cite{jacob2018quantization,bengio2013ste}.
Hardware-in-the-loop training runs the forward pass on an analog chip such as
BrainScaleS-2 and computes gradients through a software model of
it~\cite{schmitt2017hardwareloop,cramer2022surrogate}. The toolchains of digital
chips, such as Rockpool for Xylo~\cite{muir2023xylo}, model the
chip's bit-shift decays and per-step event caps during training.

%% file: Sections/9-discussion.tex
\section{Discussion and Future Work}
\label{sec:discussion}

\name targets personalized speaker identification on always-on devices with a
small local gallery, such as shared smart speakers, hearables, and home
appliances that tailor their responses to each enrolled user. The neuromorphic
chip computes the embedding of every window, and the microcontroller keeps the
gallery local and mutable, so users are enrolled and rejected without any change
to the network on the chip. Two properties bound this operating point. Rejection
becomes harder as the gallery grows, since a guest is compared against every
enrolled prototype, and the rejection EER rises from $5.5\%$ with two members to
$13.2\%$ with twenty (Sec.~\ref{sec:eval-household}). \name therefore suits
applications in which a wrong decision changes a response, and a
security-critical application would need a lower error rate or a second factor.
The chip's $32$ readout neurons are also the only interface between the two
processors, so they bound the dimension of the embedding. Because the gallery
lives on the host, it can follow its users over long deployments. The host can
decide when a new enrollment is distinct enough to accept, refresh a prototype
as a user's voice changes with age, and recalibrate the threshold table of
Sec.~\ref{sec:host-inference} as the gallery grows, all without retraining the
network.

Our future work follows two directions. The first is to follow households over
years. A speaker's voice changes as they age, and each speaker in our households
contributes a few minutes of speech, so change over years has not been
considered. We plan to collect recordings of the same households over multiple
years and use them to design host policies that refresh a member's prototype
from confidently identified decisions and recalibrate the rejection threshold as
the household changes, without retraining the network. The second direction is
hardware--algorithm co-design for future neuromorphic chips. CAST measures the
accuracy cost of each hardware constraint separately
(Table~\ref{tab:chip-loss}), which makes it a tool for chip design as well as a
training method. Before a chip is built, its design simulator can serve as the
bit-exact model of its update, and CAST can train a network through each
candidate set of event limits, weight formats, state widths, and readout widths.
We plan to choose the constraints of future chips by weighing the accuracy each
candidate reaches against its silicon area and power, and to understand when
training through hardware constraints improves generalization, so that the
limits of a chip can be designed to strengthen its models as well as to save
energy.

%% file: Sections/10-conlclusion.tex
\section{Conclusion}
\label{sec:conclusion}

We presented \name, a hybrid neuromorphic--von Neumann system for open-set speaker
identification on always-on devices, pairing a spiking embedding processor on
Xylo Audio~3 with a lightweight host for prototype scoring, enrollment, and
rejection. On $20$ Common Voice households of unseen speakers, \name reached
$99.6\%$ top-1 accuracy and $5.5\%$ rejection EER with $2$ members and
$94.9\%$ and $13.2\%$ with $20$. On $20$ further households, the deployed
chip matched or exceeded the PyTorch model it was fine-tuned from. Each
three-second window cost $0.96\,\mathrm{s}$ and $1.63\,\mathrm{mJ}$ of dynamic
energy, with $42\%$ lower latency than the fastest embedded ANN baseline and
$13.5\times$ less dynamic and $5.8\times$ less total energy than the most
energy-efficient one. These results show that neuromorphic audio hardware can
serve as the low-power embedding substrate for open-set speaker recognition at
the edge.

%% file: ref-1.bib
@article{zhang2018helloedge,
  title={Hello edge: Keyword spotting on microcontrollers},
  author={Zhang, Yundong and Suda, Naveen and Lai, Liangzhen and Chandra, Vikas},
  journal={arXiv preprint arXiv:1711.07128},
  year={2017}
}

@inproceedings{kim2021bcresnet,
  author    = {Byeonggeun Kim and Simyung Chang and Jinkyu Lee and Dooyong Sung},
  title     = {Broadcasted Residual Learning for Efficient Keyword Spotting},
  booktitle = {Proceedings of Interspeech 2021},
  pages     = {4538--4542},
  year      = {2021},
  doi       = {10.21437/Interspeech.2021-383}
}

@inproceedings{snyder2018xvectors,
  author    = {David Snyder and Daniel Garcia-Romero and Gregory Sell and Daniel Povey and Sanjeev Khudanpur},
  title     = {X-Vectors: Robust {DNN} Embeddings for Speaker Recognition},
  booktitle = {2018 IEEE International Conference on Acoustics, Speech and Signal Processing (ICASSP)},
  pages     = {5329--5333},
  publisher = {IEEE},
  year      = {2018},
  doi       = {10.1109/ICASSP.2018.8461375}
}

@inproceedings{desplanques2020ecapa,
  author    = {Brecht Desplanques and Jenthe Thienpondt and Kris Demuynck},
  title     = {{ECAPA-TDNN}: Emphasized Channel Attention, Propagation and Aggregation in {TDNN} Based Speaker Verification},
  booktitle = {Proceedings of Interspeech 2020},
  pages     = {3830--3834},
  year      = {2020},
  doi       = {10.21437/Interspeech.2020-2650}
}

@article{chen2022wavlm,
  author  = {Sanyuan Chen and Chengyi Wang and Zhengyang Chen and Yu Wu and Shujie Liu and Zhuo Chen and Jinyu Li and Naoyuki Kanda and Takuya Yoshioka and Xiong Xiao and Jian Wu and Long Zhou and Shuo Ren and Yanmin Qian and Yao Qian and Jian Wu and Michael Zeng and Xiangzhan Yu and Furu Wei},
  title   = {{WavLM}: Large-Scale Self-Supervised Pre-Training for Full Stack Speech Processing},
  journal = {IEEE Journal of Selected Topics in Signal Processing},
  volume  = {16},
  number  = {6},
  pages   = {1505--1518},
  year    = {2022},
  doi     = {10.1109/JSTSP.2022.3188113}
}

@inproceedings{wang2023campplus,
  author    = {Hui Wang and Siqi Zheng and Yafeng Chen and Luyao Cheng and Qian Chen},
  title     = {{CAM++}: A Fast and Efficient Network for Speaker Verification Using Context-Aware Masking},
  booktitle = {Proceedings of Interspeech 2023},
  pages     = {5301--5305},
  year      = {2023},
  doi       = {10.21437/Interspeech.2023-1513}
}

@inproceedings{chen2023eres2net,
  author    = {Yafeng Chen and Siqi Zheng and Hui Wang and Luyao Cheng and Qian Chen and Jiajun Qi},
  title     = {An Enhanced {Res2Net} with Local and Global Feature Fusion for Speaker Verification},
  booktitle = {Proceedings of Interspeech 2023},
  pages     = {2228--2232},
  year      = {2023},
  doi       = {10.21437/Interspeech.2023-1294}
}

@inproceedings{deng2019arcface,
  author    = {Jiankang Deng and Jia Guo and Niannan Xue and Stefanos Zafeiriou},
  title     = {{ArcFace}: Additive Angular Margin Loss for Deep Face Recognition},
  booktitle = {2019 IEEE/CVF Conference on Computer Vision and Pattern Recognition (CVPR)},
  pages     = {4690--4699},
  publisher = {IEEE},
  year      = {2019},
  doi       = {10.1109/CVPR.2019.00482}
}

@inproceedings{khosla2020supcon,
  author    = {Prannay Khosla and Piotr Teterwak and Chen Wang and Aaron Sarna and Yonglong Tian and Phillip Isola and Aaron Maschinot and Ce Liu and Dilip Krishnan},
  title     = {Supervised Contrastive Learning},
  booktitle = {Advances in Neural Information Processing Systems},
  volume    = {33},
  pages     = {18661--18673},
  year      = {2020},
  url       = {https://proceedings.neurips.cc/paper/2020/hash/d89a66c7c80a29b1bdbab0f2a1a94af8-Abstract.html}
}

@article{neftci2019surrogate,
  author  = {Emre O. Neftci and Hesham Mostafa and Friedemann Zenke},
  title   = {Surrogate Gradient Learning in Spiking Neural Networks: Bringing the Power of Gradient-Based Optimization to Spiking Neural Networks},
  journal = {IEEE Signal Processing Magazine},
  volume  = {36},
  number  = {6},
  pages   = {51--63},
  year    = {2019},
  doi     = {10.1109/MSP.2019.2931595}
}

@inbook{muir2023xylo,
  author    = {Hannah Bos and Dylan Muir},
  title     = {Sub-mW Neuromorphic {SNN} Audio Processing Applications with Rockpool and Xylo},
  booktitle = {Embedded Artificial Intelligence: Devices, Embedded Systems, and Industrial Applications},
  editor    = {Ovidiu Vermesan and Mario Diaz Nava and Bjorn Debaillie},
  series    = {River Publishers Series in Communications and Networking},
  pages     = {94--103},
  publisher = {River Publishers},
  address   = {Denmark},
  year      = {2023},
  isbn      = {9788770228213}
}

@article{bos2024micropower,
  author        = {Hannah Bos and Dylan R. Muir},
  title         = {Micro-power Spoken Keyword Spotting on Xylo Audio 2},
  journal       = {CoRR},
  volume        = {abs/2406.15112},
  year          = {2024},
  doi           = {10.48550/arXiv.2406.15112},
  url           = {https://arxiv.org/abs/2406.15112},
  eprint        = {2406.15112},
  archivePrefix = {arXiv},
  primaryClass  = {cs.NE}
}

@book{gerstner2002spiking,
  author    = {Wulfram Gerstner and Werner M. Kistler},
  title     = {Spiking Neuron Models: Single Neurons, Populations, Plasticity},
  publisher = {Cambridge University Press},
  year      = {2002},
  doi       = {10.1017/CBO9780511815706},
  isbn      = {9780521890793}
}

@inproceedings{aardila2020commonvoice,
  title={Common voice: A massively-multilingual speech corpus},
  author={Ardila, Rosana and Branson, Megan and Davis, Kelly and Kohler, Michael and Meyer, Josh and Henretty, Michael and Morais, Reuben and Saunders, Lindsay and Tyers, Francis and Weber, Gregor},
  booktitle={Proceedings of the twelfth language resources and evaluation conference},
  pages={4218--4222},
  year={2020}
}

@manual{stcubeai,
  author       = {{STMicroelectronics}},
  title        = {Getting Started with {X-CUBE-AI} Expansion Package for Artificial Intelligence ({AI})},
  organization = {STMicroelectronics},
  number       = {UM2526},
  year         = {2022},
  url          = {https://www.st.com/resource/en/user_manual/um2526-getting-started-with-xcubeai-expansion-package-for-artificial-intelligence-ai-stmicroelectronics.pdf}
}

@inproceedings{spikefit2025,
  author    = {Ivan Kartashov and Mariia Pushkareva and Iakov Karandashev},
  title     = {SpikeFit: Towards Optimal Deployment of Spiking Networks on Neuromorphic Hardware},
  booktitle = {Rethinking {AI}: Efficiency, Frugality, and Sustainability Workshop at the European Conference on Neural Information Processing Systems ({EurIPS})},
  year      = {2025},
  url       = {https://rethinking-ai.github.io/assets/papers/37_SpikeFit_Towards_Optimal_De.pdf}
}

@inproceedings{gallacher2023speakerid,
  author    = {Markus Gallacher and Carlo Alberto Boano and M. S. Arun Sankar and Utz Roedig and Willian T. Lunardi and Michael Baddeley},
  title     = {Towards Speaker Identification on Resource-Constrained Embedded Devices},
  booktitle = {Proceedings of the 21st ACM Conference on Embedded Networked Sensor Systems, SenSys 2023, Istanbul, Turkiye, November 12-17, 2023},
  pages     = {518--519},
  publisher = {ACM},
  year      = {2023},
  doi       = {10.1145/3625687.3628387},
  url       = {https://doi.org/10.1145/3625687.3628387}
}

@article{warden2018speechcommands,
  title={Speech commands: A dataset for limited-vocabulary speech recognition},
  author={Warden, Pete},
  journal={arXiv preprint arXiv:1804.03209},
  year={2018}
}

@article{david2021tflmicro,
  title={Tensorflow lite micro: Embedded machine learning for tinyml systems},
  author={David, Robert and Duke, Jared and Jain, Advait and Janapa Reddi, Vijay and Jeffries, Nat and Li, Jian and Kreeger, Nick and Nappier, Ian and Natraj, Meghna and Wang, Tiezhen and others},
  journal={Proceedings of machine learning and systems},
  volume={3},
  pages={800--811},
  year={2021}
}

@article{banbury2021mlperf,
  author  = {Colby Banbury and Vijay Janapa Reddi and Peter Torelli and Jeremy Holleman and Nat Jeffries and Csaba Kiraly and Pietro Montino and David Kanter and Sebastian Ahmed and Danilo Pau and Urmish Thakker and Antonio Torrini and Peter Warden and Jay Cordaro and Giuseppe Di Guglielmo and Javier Duarte and Stephen Gibellini and Videet Parekh and Honson Tran and Nhan Tran and Niu Wenxu and Xu Xuesong},
  title   = {{MLPerf} Tiny Benchmark},
  journal = {Proceedings of the Neural Information Processing Systems Track on Datasets and Benchmarks},
  year    = {2021},
  url     = {https://datasets-benchmarks-proceedings.neurips.cc/paper_files/paper/2021/hash/da4fb5c6e93e74d3df8527599fa62642-Abstract-round1.html}
}

@inproceedings{blouw2019benchmarking,
  author    = {Peter Blouw and Xuan Choo and Eric Hunsberger and Chris Eliasmith},
  title     = {Benchmarking Keyword Spotting Efficiency on Neuromorphic Hardware},
  booktitle = {Proceedings of the 7th Annual Neuro-inspired Computational Elements Workshop},
  pages     = {1--8},
  publisher = {ACM},
  year      = {2019},
  doi       = {10.1145/3320288.3320304},
  url       = {https://doi.org/10.1145/3320288.3320304}
}

@inproceedings{blouw2020event,
  author    = {Peter Blouw and Chris Eliasmith},
  title     = {Event-Driven Signal Processing with Neuromorphic Computing Systems},
  booktitle = {2020 IEEE International Conference on Acoustics, Speech and Signal Processing (ICASSP)},
  pages     = {8534--8538},
  publisher = {IEEE},
  year      = {2020},
  doi       = {10.1109/ICASSP40776.2020.9053043}
}

@article{reynolds2000gmmubm,
  author  = {Douglas A. Reynolds and Thomas F. Quatieri and Robert B. Dunn},
  title   = {Speaker Verification Using Adapted Gaussian Mixture Models},
  journal = {Digital Signal Processing},
  volume  = {10},
  number  = {1--3},
  pages   = {19--41},
  year    = {2000},
  doi     = {10.1006/dspr.1999.0361}
}

@article{dehak2011frontfactor,
  author  = {Najim Dehak and Patrick J. Kenny and R{\'e}da Dehak and Pierre Dumouchel and Pierre Ouellet},
  title   = {Front-End Factor Analysis for Speaker Verification},
  journal = {IEEE Transactions on Audio, Speech, and Language Processing},
  volume  = {19},
  number  = {4},
  pages   = {788--798},
  year    = {2011},
  doi     = {10.1109/TASL.2010.2064307}
}

@inproceedings{variani2014dvector,
  author    = {Ehsan Variani and Xin Lei and Erik McDermott and Ignacio Lopez Moreno and Javier Gonzalez-Dominguez},
  title     = {Deep Neural Networks for Small Footprint Text-Dependent Speaker Verification},
  booktitle = {2014 IEEE International Conference on Acoustics, Speech and Signal Processing (ICASSP)},
  pages     = {4052--4056},
  publisher = {IEEE},
  year      = {2014},
  doi       = {10.1109/ICASSP.2014.6854363}
}

@inproceedings{okabe2018attentive,
  author    = {Koji Okabe and Takafumi Koshinaka and Koichi Shinoda},
  title     = {Attentive Statistics Pooling for Deep Speaker Embedding},
  booktitle = {Proceedings of Interspeech 2018},
  pages     = {2252--2256},
  year      = {2018},
  doi       = {10.21437/Interspeech.2018-993}
}

@inproceedings{nagrani2017voxceleb,
  author    = {Arsha Nagrani and Joon Son Chung and Andrew Zisserman},
  title     = {{VoxCeleb}: A Large-Scale Speaker Identification Dataset},
  booktitle = {Proceedings of Interspeech 2017},
  pages     = {2616--2620},
  year      = {2017},
  doi       = {10.21437/Interspeech.2017-950}
}

@inproceedings{chung2018voxceleb2,
  author    = {Joon Son Chung and Arsha Nagrani and Andrew Zisserman},
  title     = {{VoxCeleb2}: Deep Speaker Recognition},
  booktitle = {Proceedings of Interspeech 2018},
  pages     = {1086--1090},
  year      = {2018},
  doi       = {10.21437/Interspeech.2018-1929}
}

@inproceedings{wan2018ge2e,
  author    = {Li Wan and Quan Wang and Alan Papir and Ignacio Lopez Moreno},
  title     = {Generalized End-to-End Loss for Speaker Verification},
  booktitle = {2018 IEEE International Conference on Acoustics, Speech and Signal Processing (ICASSP)},
  pages     = {4879--4883},
  publisher = {IEEE},
  year      = {2018},
  doi       = {10.1109/ICASSP.2018.8462665}
}

@inproceedings{snell2017prototypical,
  author    = {Jake Snell and Kevin Swersky and Richard S. Zemel},
  title     = {Prototypical Networks for Few-Shot Learning},
  booktitle = {Advances in Neural Information Processing Systems},
  volume    = {30},
  year      = {2017},
  url       = {https://proceedings.neurips.cc/paper/2017/hash/cb8da6767461f2812ae4290eac7cbc42-Abstract.html}
}

@article{scheirer2013open,
  author  = {Walter J. Scheirer and Anderson de Rezende Rocha and Archana Sapkota and Terrance E. Boult},
  title   = {Toward Open Set Recognition},
  journal = {IEEE Transactions on Pattern Analysis and Machine Intelligence},
  volume  = {35},
  number  = {7},
  pages   = {1757--1772},
  year    = {2013},
  doi     = {10.1109/TPAMI.2012.256}
}

@article{geng2021opensetsurvey,
  author  = {Chuanxing Geng and Sheng-Jun Huang and Songcan Chen},
  title   = {Recent Advances in Open Set Recognition: A Survey},
  journal = {IEEE Transactions on Pattern Analysis and Machine Intelligence},
  volume  = {43},
  number  = {10},
  pages   = {3614--3631},
  year    = {2021},
  doi     = {10.1109/TPAMI.2020.2981604}
}

@inproceedings{vaze2022openset,
  title={Open-set recognition: A good closed-set classifier is all you need},
  author={Vaze, Sagar and Han, Kai and Vedaldi, Andrea and Zisserman, Andrew},
  booktitle={International conference on learning representations},
  year={2022}
}

@article{zenke2018superspike,
  author  = {Friedemann Zenke and Surya Ganguli},
  title   = {{SuperSpike}: Supervised Learning in Multilayer Spiking Neural Networks},
  journal = {Neural Computation},
  volume  = {30},
  number  = {6},
  pages   = {1514--1541},
  year    = {2018},
  doi     = {10.1162/neco_a_01086}
}

@inproceedings{bellec2018long,
  author    = {Guillaume Bellec and Darjan Salaj and Anand Subramoney and Robert Legenstein and Wolfgang Maass},
  title     = {Long Short-Term Memory and Learning-to-Learn in Networks of Spiking Neurons},
  booktitle = {Advances in Neural Information Processing Systems},
  volume    = {31},
  year      = {2018},
  url       = {https://proceedings.neurips.cc/paper/2018/hash/c203d8a151612acf12457e4d67635a95-Abstract.html}
}

@inproceedings{shrestha2018slayer,
  author    = {Sumit Bam Shrestha and Garrick Orchard},
  title     = {{SLAYER}: Spike Layer Error Reassignment in Time},
  booktitle = {Advances in Neural Information Processing Systems},
  volume    = {31},
  year      = {2018},
  url       = {https://proceedings.neurips.cc/paper/2018/hash/82f2b308c3b01637c607ce05f52a2fed-Abstract.html}
}

@article{wu2018stbp,
  author  = {Yujie Wu and Lei Deng and Guoqi Li and Jun Zhu and Luping Shi},
  title   = {Spatio-Temporal Backpropagation for Training High-Performance Spiking Neural Networks},
  journal = {Frontiers in Neuroscience},
  volume  = {12},
  pages   = {331},
  year    = {2018},
  doi     = {10.3389/fnins.2018.00331}
}

@article{lee2020spikebackprop,
  author  = {Chankyu Lee and Syed Shakib Sarwar and Priyadarshini Panda and Gopalakrishnan Srinivasan and Kaushik Roy},
  title   = {Enabling Spike-Based Backpropagation for Training Deep Neural Network Architectures},
  journal = {Frontiers in Neuroscience},
  volume  = {14},
  pages   = {119},
  year    = {2020},
  doi     = {10.3389/fnins.2020.00119}
}

@inproceedings{rathi2020hybrid,
  author    = {Nitin Rathi and Gopalakrishnan Srinivasan and Priyadarshini Panda and Kaushik Roy},
  title     = {Enabling Deep Spiking Neural Networks with Hybrid Conversion and Spike Timing Dependent Backpropagation},
  booktitle = {International Conference on Learning Representations},
  year      = {2020},
  url       = {https://openreview.net/forum?id=B1xSperKvH}
}

@inproceedings{fang2021sewresnet,
  author    = {Wei Fang and Zhaofei Yu and Yanqi Chen and Tiejun Huang and Timoth{\'e}e Masquelier and Yonghong Tian},
  title     = {Deep Residual Learning in Spiking Neural Networks},
  booktitle = {Advances in Neural Information Processing Systems},
  volume    = {34},
  pages     = {21056--21069},
  year      = {2021},
  url       = {https://proceedings.neurips.cc/paper/2021/hash/afe434653a898da20044041262b3ac74-Abstract.html}
}

@inproceedings{deng2022tet,
  author    = {Shikuang Deng and Yuhang Li and Shanghang Zhang and Shi Gu},
  title     = {Temporal Efficient Training of Spiking Neural Network via Gradient Re-Weighting},
  booktitle = {International Conference on Learning Representations},
  year      = {2022},
  url       = {https://openreview.net/forum?id=_XNtisL32jv}
}

@inproceedings{zhou2023spikformer,
  title={Spikformer: When spiking neural network meets transformer},
  author={Zhou, Zhaokun and Zhu, Yuesheng and He, Chao and Wang, Yaowei and Yan, Shuicheng and Tian, Yonghong and Yuan, Li},
  booktitle={International Conference on Learning Representations},
  year={2023}
}

@inproceedings{zhou2024qkformer,
  author    = {Chenlin Zhou and Han Zhang and Zhaokun Zhou and Liutao Yu and Liwei Huang and Xiaopeng Fan and Li Yuan and Zhengyu Ma and Huihui Zhou and Yonghong Tian},
  title     = {{QKFormer}: Hierarchical Spiking Transformer Using {Q-K} Attention},
  booktitle = {Advances in Neural Information Processing Systems},
  volume    = {37},
  year      = {2024},
  url       = {https://proceedings.neurips.cc/paper_files/paper/2024/hash/179f5dcdeedc149443ebd3ba70811dbd-Abstract-Conference.html}
}

@article{roy2019spikebased,
  author  = {Kaushik Roy and Akhilesh Jaiswal and Priyadarshini Panda},
  title   = {Towards Spike-Based Machine Intelligence with Neuromorphic Computing},
  journal = {Nature},
  volume  = {575},
  number  = {7784},
  pages   = {607--617},
  year    = {2019},
  doi     = {10.1038/s41586-019-1677-2}
}

@article{akopyan2015truenorth,
  author  = {Filipp Akopyan and Jun Sawada and Andrew Cassidy and Rodrigo Alvarez-Icaza and John Arthur and Paul Merolla and Nabil Imam and Yutaka Nakamura and Pallab Datta and Gi-Joon Nam and Brian Taba and Michael Beakes and Bernard Brezzo and Jente B. Kuang and Rajit Manohar and William P. Risk and Bryan Jackson and Dharmendra S. Modha},
  title   = {{TrueNorth}: Design and Tool Flow of a 65 {mW} 1 Million Neuron Programmable Neurosynaptic Chip},
  journal = {IEEE Transactions on Computer-Aided Design of Integrated Circuits and Systems},
  volume  = {34},
  number  = {10},
  pages   = {1537--1557},
  year    = {2015},
  doi     = {10.1109/TCAD.2015.2474396}
}

@article{davies2018loihi,
  author  = {Mike Davies and Narayan Srinivasa and Tsung-Han Lin and Gautham Chinya and Yongqiang Cao and Sri Harsha Choday and Georgios Dimou and Prasad Joshi and Nabil Imam and Shweta Jain and Yuyun Liao and Chit-Kwan Lin and Andrew Lines and Ruokun Liu and Deepak Mathaikutty and Steve McCoy and Arnab Paul and Jonathan Tse and Guruguhanathan Venkataramanan and Yi-Hsin Weng and Andreas Wild and Yoonseok Yang and Hong Wang},
  title   = {Loihi: A Neuromorphic Manycore Processor with On-Chip Learning},
  journal = {IEEE Micro},
  volume  = {38},
  number  = {1},
  pages   = {82--99},
  year    = {2018},
  doi     = {10.1109/MM.2018.112130359}
}

@article{furber2014spinnaker,
  author  = {Steve B. Furber and Francesco Galluppi and Steve Temple and Luis A. Plana},
  title   = {The {SpiNNaker} Project},
  journal = {Proceedings of the IEEE},
  volume  = {102},
  number  = {5},
  pages   = {652--665},
  year    = {2014},
  doi     = {10.1109/JPROC.2014.2304638}
}

@article{moradi2018dynapse,
  author  = {Saber Moradi and Ning Qiao and Fabio Stefanini and Giacomo Indiveri},
  title   = {A Scalable Multicore Architecture With Heterogeneous Memory Structures for Dynamic Neuromorphic Asynchronous Processors ({DYNAPs})},
  journal = {IEEE Transactions on Biomedical Circuits and Systems},
  volume  = {12},
  number  = {1},
  pages   = {106--122},
  year    = {2018},
  doi     = {10.1109/TBCAS.2017.2759700}
}

@article{davies2021loihi,
  author  = {Mike Davies and Andreas Wild and Garrick Orchard and Yulia Sandamirskaya and Gabriel A. Fonseca Guerra and Prasad Joshi and Philipp Plank and Sumedh R. Risbud},
  title   = {Advancing Neuromorphic Computing With Loihi: A Survey of Results and Outlook},
  journal = {Proceedings of the IEEE},
  volume  = {109},
  number  = {5},
  pages   = {911--934},
  year    = {2021},
  doi     = {10.1109/JPROC.2021.3067593}
}

@article{yik2025neurobench,
  author  = {Yik, Jason and Van den Berghe, Korneel and den Blanken, Douwe and Bouhadjar, Younes and Fabre, Maxime and Hueber, Paul and Ke, Weijie and Khoei, Mina A. and Kleyko, Denis and Pacik-Nelson, Noah and Pierro, Alessandro and Stratmann, Philipp and Sun, Pao-Sheng Vincent and Tang, Guangzhi and Wang, Shenqi and Zhou, Biyan and Ahmed, Soikat Hasan and Joseph, George Vathakkattil and Leto, Benedetto and Micheli, Aurora and Mishra, Anurag Kumar and Lenz, Gregor and Sun, Tao and Ahmed, Zergham and Akl, Mahmoud and Anderson, Brian and Andreou, Andreas G. and Bartolozzi, Chiara and Basu, Arindam and Bogdan, Petrut and Bohte, Sander and Buckley, Sonia and Cauwenberghs, Gert and Chicca, Elisabetta and Corradi, Federico and de Croon, Guido and Danielescu, Andreea and Daram, Anurag and Davies, Mike and Demirag, Yigit and Eshraghian, Jason and Fischer, Tobias and Forest, Jeremy and Fra, Vittorio and Furber, Steve and Furlong, P. Michael and Gilpin, William and Gilra, Aditya and Gonzalez, Hector A. and Indiveri, Giacomo and Joshi, Siddharth and Karia, Vedant and Khacef, Lyes and Knight, James C. and Kriener, Laura and Kubendran, Rajkumar and Kudithipudi, Dhireesha and Liu, Shih-Chii and Liu, Yao-Hong and Ma, Haoyuan and Manohar, Rajit and Margarit-Taul{\'e}, Josep Maria and Mayr, Christian and Michmizos, Konstantinos and Muir, Dylan R. and Neftci, Emre and Nowotny, Thomas and Ottati, Fabrizio and Ozcelikkale, Ayca and Panda, Priyadarshini and Park, Jongkil and Payvand, Melika and Pehle, Christian and Petrovici, Mihai A. and Posch, Christoph and Renner, Alpha and Sandamirskaya, Yulia and Schaefer, Clemens J. S. and van Schaik, Andr{\'e} and Schemmel, Johannes and Schmidgall, Samuel and Schuman, Catherine and Seo, Jae-sun and Sheik, Sadique and Shrestha, Sumit Bam and Sifalakis, Manolis and Sironi, Amos and Stewart, Kenneth and Stewart, Matthew and Stewart, Terrence C. and Timcheck, Jonathan and T{\"o}men, Nergis and Urgese, Gianvito and Verhelst, Marian and Vineyard, Craig M. and Vogginger, Bernhard and Yousefzadeh, Amirreza and Zohora, Fatima Tuz and Frenkel, Charlotte and Reddi, Vijay Janapa},
  title   = {The neurobench framework for benchmarking neuromorphic computing algorithms and systems},
  journal = {Nature Communications},
  volume  = {16},
  pages   = {1545},
  year    = {2025},
  doi     = {10.1038/s41467-025-56739-4}
}

@inproceedings{xie2019utterance,
  author    = {Xie, Weidi and Nagrani, Arsha and Chung, Joon Son and Zisserman, Andrew},
  title     = {Utterance-Level Aggregation for Speaker Recognition in the Wild},
  booktitle = {Proceedings of the IEEE International Conference on Acoustics, Speech and Signal Processing (ICASSP)},
  pages     = {5791--5795},
  year      = {2019},
  doi       = {10.1109/ICASSP.2019.8683120},
  eprint    = {1902.10107},
  archivePrefix = {arXiv},
  url       = {https://arxiv.org/abs/1902.10107}
}

@misc{st2024stm32l476ds,
  author       = {{STMicroelectronics}},
  title        = {{STM32L476xx ultra-low-power Arm Cortex-M4 32-bit MCU datasheet}},
  howpublished = {DS10198 Rev. 11},
  year         = {2024},
  url          = {https://www.st.com/resource/en/datasheet/stm32l476je.pdf},
  note         = {Accessed 2026-06-03}
}

@misc{st2024stm32u575ds,
  author       = {{STMicroelectronics}},
  title        = {{STM32U575xx ultra-low-power Arm Cortex-M33 32-bit MCU datasheet}},
  howpublished = {DS13737},
  year         = {2024},
  url          = {https://www.st.com/resource/en/datasheet/stm32u575ag.pdf},
  note         = {Accessed 2026-10-28}
}

@article{vandermaaten2008tsne,
  title   = {Visualizing data using t-{SNE}},
  author  = {van der Maaten, Laurens and Hinton, Geoffrey},
  journal = {Journal of Machine Learning Research},
  volume  = {9},
  number  = {86},
  pages   = {2579--2605},
  year    = {2008}
}

@misc{synsense2023xyloaudio3ds,
  author       = {{SynSense}},
  title        = {{Xylo Audio 3 Customer Datasheet}},
  howpublished = {Customer datasheet, August 2023},
  year         = {2023},
  url          = {https://www.synsense.ai/wp-content/uploads/2025/02/XyloAudio-3-Customer-Datasheet-2023.08.pdf},
  note         = {Accessed 2026-06-04}
}

@inproceedings{balian2021smallfootprint,
  author    = {Julien Balian and Raffaele Tavarone and Mathieu Poumeyrol and Alice Coucke},
  title     = {Small Footprint Text-Independent Speaker Verification for Embedded Systems},
  booktitle = {ICASSP 2021 - 2021 IEEE International Conference on Acoustics, Speech and Signal Processing (ICASSP)},
  pages     = {6179--6183},
  year      = {2021},
  eprint    = {2011.01709},
  archivePrefix = {arXiv},
  primaryClass  = {eess.AS},
  url       = {https://arxiv.org/abs/2011.01709}
}

@inproceedings{tan2021sharedDevices,
  author    = {Zhenning Tan and Yuguang Yang and Eunjung Han and Andreas Stolcke},
  title     = {Improving Speaker Identification for Shared Devices by Adapting Embeddings to Speaker Subsets},
  booktitle = {Proceedings of IEEE ASRU},
  pages     = {1124--1131},
  year      = {2021},
  doi       = {10.1109/ASRU51503.2021.9687975}
}

@inproceedings{kc2022openfeat,
  author    = {Kishan K C and Zhenning Tan and Long Chen and Minho Jin and Eunjung Han and Andreas Stolcke and Chul Lee},
  title     = {{OpenFEAT}: Improving Speaker Identification by Open-Set Few-Shot Embedding Adaptation with Transformer},
  booktitle = {Proceedings of IEEE ICASSP},
  pages     = {7062--7066},
  year      = {2022},
  doi       = {10.1109/ICASSP43922.2022.9746709}
}

@inproceedings{liu2020peeler,
  author    = {Bo Liu and Hao Kang and Haoxiang Li and Gang Hua and Nuno Vasconcelos},
  title     = {Few-Shot Open-Set Recognition Using Meta-Learning},
  booktitle = {Proceedings of the IEEE/CVF Conference on Computer Vision and Pattern Recognition},
  pages     = {8798--8807},
  year      = {2020}
}

@article{zhu2017prune,
  title   = {To Prune, or Not to Prune: Exploring the Efficacy of Pruning for Model Compression},
  author  = {Zhu, Michael and Gupta, Suyog},
  journal = {arXiv preprint arXiv:1710.01878},
  year    = {2017}
}

@inproceedings{jacob2018quantization,
  title     = {Quantization and Training of Neural Networks for Efficient Integer-Arithmetic-Only Inference},
  author    = {Jacob, Benoit and Kligys, Skirmantas and Chen, Bo and Zhu, Menglong and Tang, Matthew and Howard, Andrew and Adam, Hartwig and Kalenichenko, Dmitry},
  booktitle = {Proceedings of the IEEE Conference on Computer Vision and Pattern Recognition (CVPR)},
  pages     = {2704--2713},
  year      = {2018}
}

@article{bengio2013ste,
  title   = {Estimating or Propagating Gradients Through Stochastic Neurons for Conditional Computation},
  author  = {Bengio, Yoshua and L{\'e}onard, Nicholas and Courville, Aaron},
  journal = {arXiv preprint arXiv:1308.3432},
  year    = {2013}
}

@article{pearson1901lines,
  author  = {Karl Pearson},
  title   = {On Lines and Planes of Closest Fit to Systems of Points in Space},
  journal = {The London, Edinburgh, and Dublin Philosophical Magazine and Journal of Science},
  volume  = {2},
  number  = {11},
  pages   = {559--572},
  year    = {1901},
  doi     = {10.1080/14786440109462720}
}

@article{fisher1936use,
  author  = {Ronald A. Fisher},
  title   = {The Use of Multiple Measurements in Taxonomic Problems},
  journal = {Annals of Eugenics},
  volume  = {7},
  number  = {2},
  pages   = {179--188},
  year    = {1936},
  doi     = {10.1111/j.1469-1809.1936.tb02137.x}
}

@inproceedings{schmitt2017hardwareloop,
  title     = {Neuromorphic Hardware in the Loop: Training a Deep Spiking Network on the {BrainScaleS} Wafer-Scale System},
  author    = {Schmitt, Sebastian and Kl{\"a}hn, Johann and Bellec, Guillaume and Gr{\"u}bl, Andreas and G{\"u}ttler, Maurice and Hartel, Andreas and Hartmann, Stephan and Husmann, Dan and Husmann, Kai and Jeltsch, Sebastian and others},
  booktitle = {International Joint Conference on Neural Networks (IJCNN)},
  pages     = {2227--2234},
  year      = {2017},
  doi       = {10.1109/IJCNN.2017.7966125}
}

@article{cramer2022surrogate,
  title   = {Surrogate Gradients for Analog Neuromorphic Computing},
  author  = {Cramer, Benjamin and Billaudelle, Sebastian and Kanya, Simeon and Leibfried, Aron and Gr{\"u}bl, Andreas and Karasenko, Vitali and Pehle, Christian and Schreiber, Korbinian and Stradmann, Yannik and Weis, Johannes and Schemmel, Johannes and Zenke, Friedemann},
  journal = {Proceedings of the National Academy of Sciences},
  volume  = {119},
  number  = {4},
  pages   = {e2109194119},
  year    = {2022},
  doi     = {10.1073/pnas.2109194119}
}
